\documentclass[conference]{IEEEtran}
\IEEEoverridecommandlockouts
\usepackage{cite}
\usepackage{amssymb,amsfonts}
\usepackage{amsmath}
\usepackage{algorithmic}
\usepackage{graphicx}
\usepackage{textcomp}
\usepackage{xcolor}
\usepackage{bm}
\usepackage{url}
\usepackage{multirow}
\usepackage{marvosym}
\usepackage{pifont}
\usepackage[font=footnotesize,labelfont=rm,textfont=rm]{subcaption}

\def\BibTeX{{\rm B\kern-.05em{\sc i\kern-.025em b}\kern-.08em
    T\kern-.1667em\lower.7ex\hbox{E}\kern-.125emX}}
\begin{document}

\newcommand{\pch}[1]{{\color{black}{#1}}}
\newcommand{\del}[1]{{\color{red}{#1}}}

\title{TimeSeriesBench: An Industrial-Grade Benchmark for Time Series Anomaly Detection Models
}

\author{
\IEEEauthorblockN{
Haotian Si\IEEEauthorrefmark{2}\IEEEauthorrefmark{1}, 
Jianhui Li\IEEEauthorrefmark{2}\textsuperscript{\Letter}\thanks{\Letter ~Corresponding author.},
%\textsuperscript{\Letter},
Changhua Pei\IEEEauthorrefmark{2}\IEEEauthorrefmark{6},
Hang Cui\IEEEauthorrefmark{3},
Jingwen Yang\IEEEauthorrefmark{2}\IEEEauthorrefmark{1},
Yongqian Sun\IEEEauthorrefmark{4}
},
\IEEEauthorblockN{
Shenglin Zhang\IEEEauthorrefmark{4},
Jingjing Li\IEEEauthorrefmark{2},
Haiming Zhang\IEEEauthorrefmark{2},
Jing Han\IEEEauthorrefmark{7},
Dan Pei\IEEEauthorrefmark{5},
Gaogang Xie\IEEEauthorrefmark{2}
}

\IEEEauthorblockA{
\IEEEauthorrefmark{2} Computer Network Information Center, CAS, Beijing, China,
\IEEEauthorrefmark{4} Nankai University, Tianjin, China}
\IEEEauthorblockA{
\IEEEauthorrefmark{1} University of Chinese Academy of Sciences, Beijing, China,
\IEEEauthorrefmark{3} Jilin University, Changchun, China}
\IEEEauthorblockA{
\IEEEauthorrefmark{6} Hangzhou Institute for Advanced Study, University of Chinese Academy of Sciences, Hangzhou, China
}
\IEEEauthorblockA{
\IEEEauthorrefmark{5} Tsinghua University, Beijing, China,
\IEEEauthorrefmark{7} ZTE Co., Ltd, China}
\IEEEauthorblockA{\{htsi, lijh, chpei\}@cnic.cn}
% \IEEEauthorblockA{jimmy@cstnet.cn, diaozulong@ict.ac.cn, peidan@tsinghua.edu.cn}
}

\maketitle

\begin{abstract}
\pch{Time series anomaly detection (TSAD) has gained significant attention due to its real-world applications to improve the stability of modern software systems. However, there is no effective way to verify whether they can meet the requirements for real-world deployment.
Firstly, current algorithms typically train a specific model for each time series. Maintaining such many models is impractical in a large-scale system with tens of thousands of curves. The performance of using merely one unified model to detect anomalies remains unknown. Secondly, most TSAD models are trained on the historical part of a time series and are tested on its future segment. In distributed systems, however, there are frequent system deployments and upgrades, with new, previously unseen time series emerging daily. The performance of testing newly incoming unseen time series on current TSAD algorithms remains unknown. Lastly, the assumptions of the evaluation metrics in existing benchmarks are far from practical demands. To solve the above-mentioned problems, we propose an industrial-grade benchmark TimeSeriesBench. We assess the performance of existing algorithms across more than 168 evaluation settings and provide comprehensive analysis for the future design of anomaly detection algorithms. An industrial dataset is also released along with TimeSeriesBench.}
\end{abstract}


\section{Introduction}

% The rapid expansion of cloud-based applications and the Web of Things (WoT) has resulted in a deluge of time-series data across a wide array of sectors, including but not limited to cloud systems~\cite{cloud1, cloud2, cloud3}, Web traffic~\cite{webnet1},  cybersecurity~\cite{cyber1}, and healthcare~\cite{health1, health2}. 

In recent years, we have witnessed the rapid growth in the scale of software systems and services.
To ensure service quality assurance and maintain user satisfaction, IT operation engineers must monitor the time series metrics like response time or success rate to judge if there are system failures. Time series anomaly detection (TSAD) aims to identify irregular patterns in these time series data, which indicate potential or existing system failures. The outliers can often signify critical deviations from the norm, such as impending machine failures in industrial processes~\cite{omnipa3} and malignant network traffic attacks in web applications~\cite{nettra}. Accurate and timely anomaly detection can help to reduce the mean time to repair, reduce the loss in revenue and maintain the reputation and branding for a company. 

Owing to the clear significance and applicability of TSAD in these real-world scenarios, in recent years, a variety of anomaly detection methods, particularly those based on deep learning, are burgeoning incessantly~\cite{anotrans, donut, model1, model2, model3, model4, cad, icse22}. Nonetheless, there is a considerable divergence across the results reported by different papers even on the same dataset, as they employed various evaluation criteria and learning schemas. Despite the existence of some benchmarks for anomaly detection algorithms~\cite{bench1, bench2, suite_tods, suite_exa, suite_tsb, bench3}, they struggle to provide practical guidance for industrial domain experts to apply and develop deep learning methods as they pay more attention to statistical methods. In general, applying TSAD algorithms to practical systems faces these main obstacles:
\textbf{(\uppercase\expandafter{\romannumeral1})} Training specific model for each time series and deploying an exclusive model for each time series results in unaffordable maintenance costs. 
\textbf{(\uppercase\expandafter{\romannumeral2})} New systems are more likely to encounter issues when first launched or upgraded, which makes anomaly detection even more essential despite their almost non-existent volume of historical data.
% \textbf{(\uppercase\expandafter{\romannumeral3})} In different systems, administrators have varying focal points for anomaly detection. In some scenarios, administrators prioritize the prompt detection of anomalies (delay-k); in others, they require the identification of every point within a continuous anomaly sequence (point-adjustment). In yet other contexts, administrators may only need to detect the onset of a continuous anomaly once (as with our improved xxx metric). Any single evaluation method is inherently biased.
\textbf{(\uppercase\expandafter{\romannumeral3})} Existing metrics hold one-sided assumptions for evaluating how well the algorithms perform, which cannot offer an effective reference for industrial practice. 
\textbf{(\uppercase\expandafter{\romannumeral4})} There is no platform continuously integrating new methods and comparing them in a unified and intuitive manner, similar to the GLUE Leaderboard in NLP~\cite{glue}, which prevents experts in the industry from keeping pace with the latest algorithms and advancements in academia.

\begin{figure*}[tbp]
    \centering
    \includegraphics[width=0.9\textwidth]{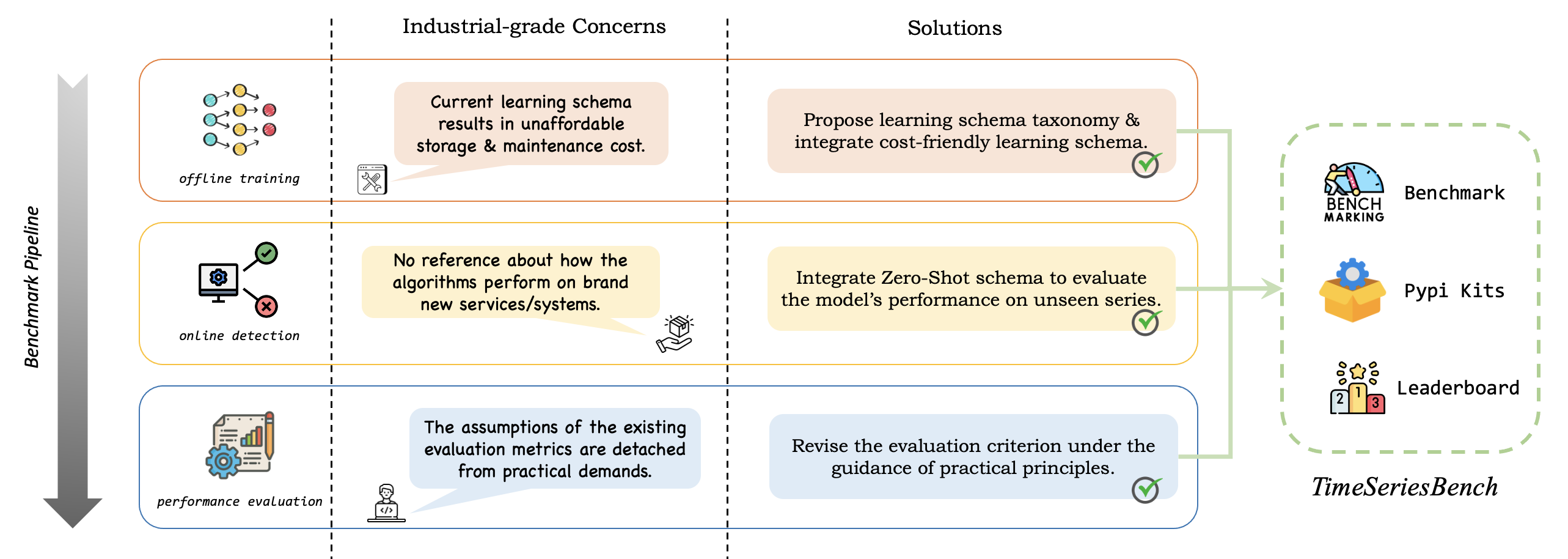}
    \caption{The industrial-grade concerns and TimeSeriesBench's solutions on benchmarking univariate time series anomaly detection algorithms.}
    \label{Fig.overview}
\end{figure*}

In response to the four challenges mentioned earlier, we propose \textbf{TimeSeriesBench}, an industrial-grade benchmark for evaluating time series anomaly detection. This benchmark has four main features:
\textbf{(\uppercase\expandafter{\romannumeral1})} To address the issue of high maintenance costs and non-deployability caused by existing works requiring a separate model for each time series, we adopt an All-in-One training paradigm to assess the detection performance of current algorithms when only one unified model is trained.
\textbf{(\uppercase\expandafter{\romannumeral2})} To tackle the inability of existing works to handle new time series, we employ a Zero-Shot inference paradigm during evaluation. By using a novel data-splitting method, we assess the model's detection performance on previously unseen curves without retraining or fine-tuning new time series.
\textbf{(\uppercase\expandafter{\romannumeral3})} We thoroughly integrate existing evaluation criteria, and propose our own event-based evaluation metric to ensure that benchmark results can provide effective guidance for industrial deployment. Also, TimeSeriesBench provides a series of well-known metrics, which can meet the detection needs of different downstream application scenarios.
% Whether it is point anomalies, interval anomalies, or the timeliness of anomaly detection, appropriate evaluation metrics can be found on TimeSeriesBench.
\textbf{(\uppercase\expandafter{\romannumeral4})} To solve the problem of industry experts not having access to a unified and continuously updated set of evaluation results, we have built an online leaderboard. This leaderboard combines the above three features and conducts a comprehensive evaluation across more than 168 evaluation settings in terms of training and testing paradigms, evaluation metrics, and multiple datasets. 

Based on TimeSeriesBench, we evaluate several representative time series anomaly detection methods ranging from statistical machine learning and deep learning to large time series models. From this, we have made some novel observations, conducted detailed analyses, and offered insights for algorithm design and deployment. To name a few, models that employ variational autoencoder exhibit good detection performance for pattern-wise anomalies, and so far the general time series models struggle to outperform methods specifically designed for anomaly detection tasks.

The paper's main contributions are as follows:
\begin{itemize}
    \item We present the first online leaderboard for time series anomaly detection algorithms, which upgrades the existing evaluation frameworks across multiple dimensions such as training, inference, evaluation and datasets. It effectively supports the industry experts' need to select the best academic algorithms and provides an industrial-grade evaluation method for academic algorithms, ensuring the deployability of future algorithms.

    \item \pch{For the first time, we employ the cost-friendly all-in-one and zero-shot settings to evaluate several well-known state-of-the-art (SOTA) anomaly detection methods on TimeSeriesBench and discover some enlightening conclusions, providing directions for future optimization.}

    \item We develop and publish a comprehensive evaluation toolkit built with Python named EasyTSAD, providing a one-stop solution for data processing, model training, and assessment, which we have made open source to the community to accelerate the efficiency of existing anomaly detection algorithm optimizations.

    \item To address the issue of inaccurate anomaly labeling in existing public datasets~\cite{dataflaw}, we collaborated with a global large company and invited business system experts to meticulously annotate anomalies in the online system. After detailed calibration, we have also made the dataset publicly available to the community as part of TimeSeriesBench, supplementing existing anomaly detection datasets. The codes, data and the online leaderboard are released publicly\footnote{The code is available at https://github.com/CSTCloudOps/EasyTSAD. The new proposed dataset is available at https://github.com/CSTCloudOps/Dataset-for-TSAD, and the online leaderboard is available at https://adeval.cstcloud.cn.}.
\end{itemize}

\section{PRELIMINARIES}

\subsection{Time Series Anomaly Detection}
A time series consists of successive observations of a metric (e.g., queries per second, sensor value, etc.) over a long period of time. The series can be represented as $T={x_1, x_2, \cdots, x_n}$, where $x_i$ represents an observation at timestamp $i$, and $n$ denotes the length of series. An anomaly within these time series data can be identified as a single point or a sequence of points that diverge significantly from their former customary patterns observed in the sequence. Anomaly detection methods project sequence observations into a probability distribution space to represent the degree of anomaly of the current observation, namely \textit{anomaly score}, and compare it with a predefined threshold to determine whether an anomaly has occurred. Due to the scarcity of anomaly labels, existing deep-learning methods often opt for self-supervised training approaches.

\subsection{Anomaly Types}
\begin{figure}[htb]
    \centering
	\subfloat[point-wise outlier]{\includegraphics[width=.4\linewidth]{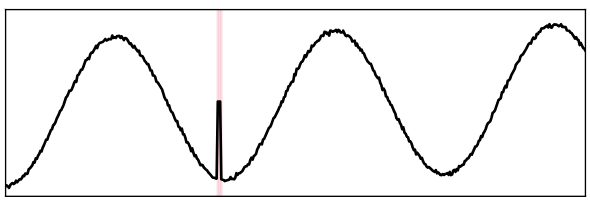}}\hspace{20pt}
    \subfloat[pattern-wise outlier]{\includegraphics[width=.4\linewidth]{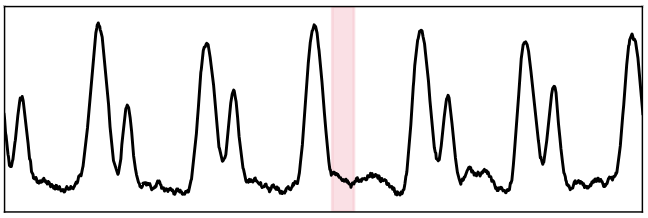}}
    \caption{Illustrations of anomaly types. Anomalous segments are highlighted in pink.}
    \label{fig.ano_type}
\end{figure}

Following the behavior-driven taxonomy~\cite{suite_tods}, anomaly types can be roughly divided into point-wise outliers and pattern-wise outliers (Fig.~\ref{fig.ano_type}). Point-wise outliers denote unexpected incidents like spikes or glitches on individual time points or within very short periods of time. Pattern-wise outliers represent anomalous subsequences that span over a certain period of time, often characterized as discordances or inconsistencies within the data.

\section{TimeSeriesBench Settings}
We launch TimeSeriesBench to integrate various aspects that are widely concerned into a pipeline to meet the diverse needs of scientific research and practical applications, and further promote the development of communities related to the field of time series anomaly detection. Aside from model implementation, the platform also takes some controversial issues, for instance, learning schema and evaluation criteria into consideration, to provide the researchers and engineers with a more comprehensive view of whether the models perform well in specific scenarios. In this section, we will illustrate the motivation and the implementation details of each inductive setting.

\subsection{Datasets}
High-quality datasets are the prerequisite for effective training and reasonable performance evaluations for models. Unfortunately, several publicly available time series datasets are claimed flawed due to their unrealistic anomaly density, or mislabeled ground truth~\cite{dataflaw}. Moreover, the proportion of different types of outliers can vary widely among datasets as the time series are gathered from irrelevant domains and applications. For example, Yahoo~\cite{yahoo} is dominated by global and contextual point-wise outliers on the basis of the behavior-driven taxonomy proposed by ~\cite{suite_tods} (the trivial first-order difference can achieve a comparable performance), while pattern-wise outlier plays a significant role in UCR archive~\cite{UCR} (shown in Sec.~\ref{sec:result}). To this end, we gather multiple well-labeled real-world datasets covering many application domains to diversify the distributions of anomalies, meanwhile excluding markedly flawed datasets (some cases in ~\cite{dataflaw}). Furthermore, we introduce a synthetic dataset generated from TODS~\cite{suite_tods} for the convenience of specific case analysis due to its good interpretability, because the time series adheres to well-designed distributions. 

\noindent \textbf{Real-world datasets.} We employ AIOPS~\cite{KPI}, WSD~\cite{WSD}, Yahoo~\cite{yahoo}, NAB~\cite{NAB}, and UCR~\cite{UCR} for evaluations. Real-world datasets are more susceptible to uncertainties and tend to mix up with even inexplicable noise, which raises the demand in terms of model robustness. As the classes of anomalies are imbalanced among these datasets, practitioners focusing on the general-purpose model should refer to detection performance on \textit{all} datasets, while application-oriented tasks need to pay more attention to performance on datasets that are strikingly similar to specific scenes. Meanwhile, we release a new dataset collected from the production environment, called NEK (Network Equipment KPI), for a more comprehensive evaluation.

\noindent \textbf{Synthetic dataset.} TODS~\cite{suite_tods} introduces a novel taxonomy, categorizes anomalies into five types, and publishes an anomaly generation toolkit in line with their claims. We conduct modifications based on their codes to generate longer and non-trivial anomalies, which would allow researchers to determine if models are capable of particular types of anomalies in a straightforward manner.

\subsection{Learning Schema}
Existing benchmarks take it for granted that the training process of the detector should comply with the naive task-specific schema, more specifically, training an exclusive detector for each time series solely leveraging its own historical patterns. We claim that it's time to break the prevailing stereotype, especially for deep learning models, given the great transformations observed in application scenarios during recent years. On the one hand, with the maturation and improvement of surveillance systems in various fields, the quantity of time series to be monitored has experienced a substantial increase, reaching several orders of magnitude higher than before. This significantly raised the storage overheads if detectors are exclusive. On the other hand, the previous years witnessed many seminal works pertinent to foundational models for time series forecasting or other general tempo tasks~\cite{tempo, timellm, timegpt, gpt4ts}. We aspire to establish a foundation for the standardized training and evaluation of large-scale, general-purpose time series models, especially those designed for anomaly detection. Thus we introduce two novel learning schemas, called all-in-one mode and zero-shot mode, to assess the performance variations of models in scenarios involving large-scale data or zero-shot settings.

\noindent\textbf{Naive schema.} In this mode, we input a single time series for model training/fitting, and the trained detector is specifically employed for online detection on that particular series. Intuitively, this facilitates the model to make a more precise description of the temporal pattern given enough data. Nevertheless, it is worth noting that the volume of a single series tends to be insufficient.

\noindent\textbf{All-in-one schema.} In this mode, only one unified model instance is trained using all the sequences in the dataset, and the trained model is then applied for real-time anomaly detection across all sequences within the dataset. This schema exposes more patterns embedded in various series to the model, thereby providing an additional opportunity for the model to learn common and inherent traits shared among the time series. However, due to the differing definitions of anomalies across series, this carries the risk of confounding the model with conflicting information when detecting anomalies online (i.e., an anomaly present in a specific curve may not necessarily be considered an anomaly in other curves). Several novel methods~\cite{tfad, srcnn} have adopted this schema driven by practical needs.

\noindent\textbf{Zero-shot schema.} In zero-shot mode, the whole dataset is split into two disjoint subsets. One subset is employed for model training and the other is used to evaluate the detection performance. This schema has been devised in light of practical considerations. Specifically, it addresses situations where a system is newly deployed with no prior historical data and a robust and adaptable model is required to navigate through this gap. This places higher demands on the model's ability to capture the intrinsic representations of time series.

\begin{figure}[tb]
    \centering
    \subfloat[naive]{\includegraphics[width=0.23\textwidth]{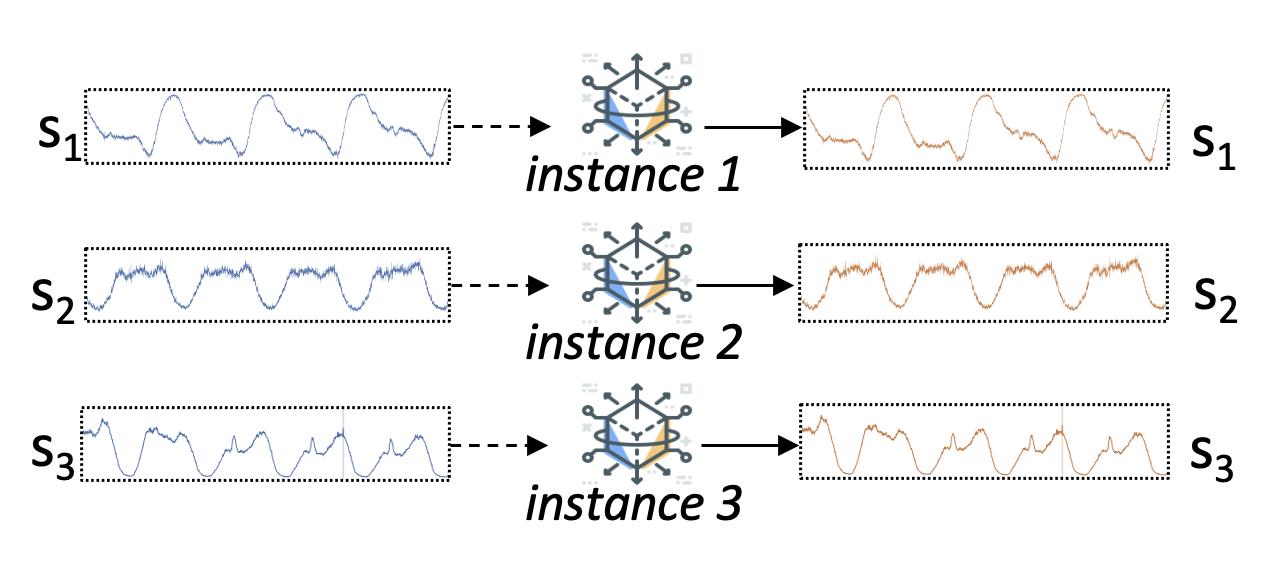}}
    \hspace{5pt}%
    \subfloat[all-in-one]{%
        \includegraphics[width=0.23\textwidth]{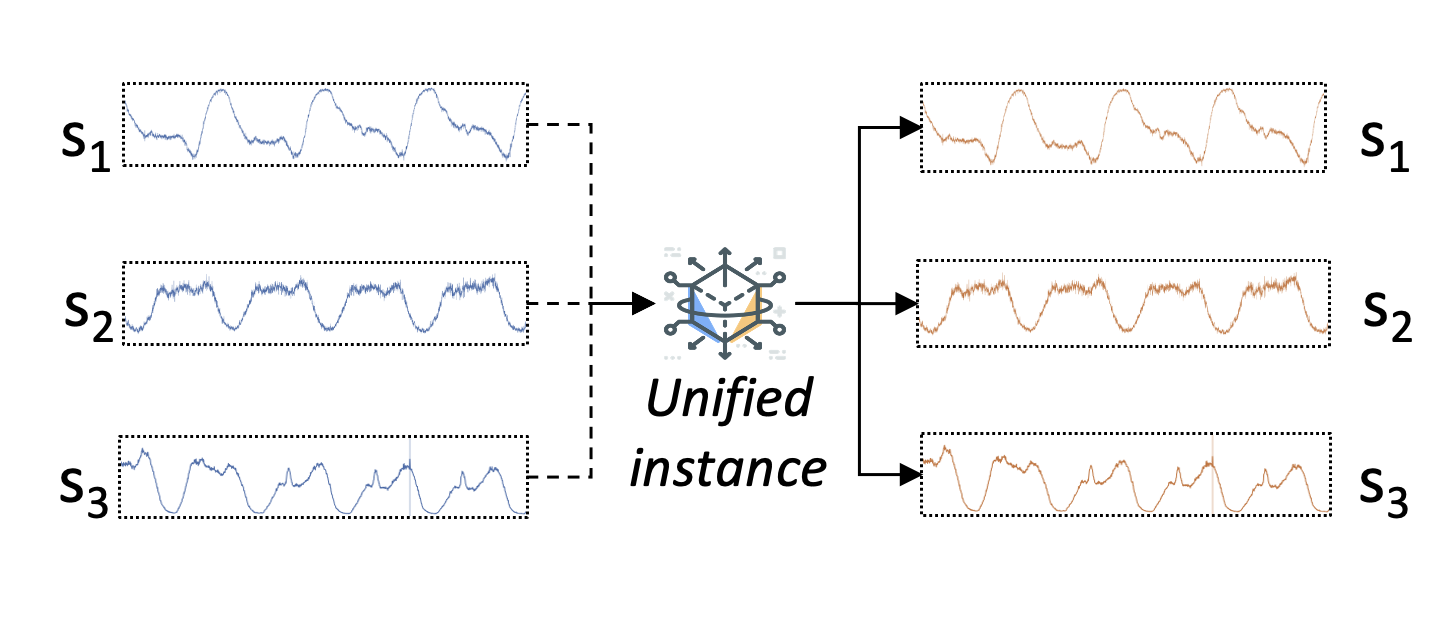}%
    }
    \\ \vspace{5pt}
    \subfloat[zero-shot]{%
        \includegraphics[width=0.48\textwidth]{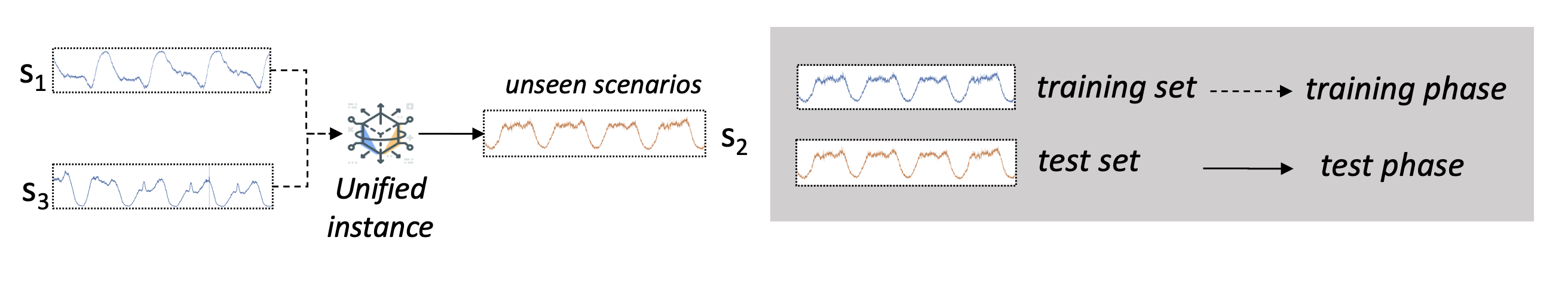}%
    }
    
    \caption{Illustrations of three learning schemas. \protect$s_n$ indicates the n-th time series in the dataset.}
    \label{fig:mode}
\end{figure}
% \begin{figure}[htb]
% 	\centering
% 	\subfloat[one-by-one]{\includegraphics[width=.29\linewidth]{figs/obomod.png}}\hspace{2pt} \tikz[overlay]\draw[dashed] (0,-0pt) -- (0,50pt);\hspace{2pt}
%     \subfloat[all-in-one]{\includegraphics[width=.3\linewidth]{figs/aiomod.png}}\hspace{2pt} \tikz[overlay]\draw[dashed] (0,-0pt) -- (0,50pt);\hspace{2pt}
%     \subfloat[zero-shot]{\includegraphics[width=.3\linewidth]{figs/transmod.png}}
% 	\caption{Illustrations of three learning schemas. $s_n$ indicates the n-th time series in the dataset.}
%     \label{fig.mode}
% \end{figure}

\subsection{Evaluation Criteria}
\label{sec:eva_cri}

A flawless evaluation criteria serves as the foundation for not only assessing the effectiveness of methods but also guiding model parameters optimization. Nevertheless, recent research has uncovered substantial limitations in commonly employed evaluation criteria, on which some seemingly absurd methods (e.g. noise generated by Random Guess) can outperform all others~\cite{evaflaw1tisat, evaflaw2arxiv, evaflaw3aaai, eventf1}. Also, some assumptions employed by certain evaluation criteria also conflict with the principles of industrial practice. Therefore, it is imperative to revise an anti-cheating and industrial-oriented evaluation criterion for our benchmark. \textit{It is crucial to emphasize that the selection of criterion is closely tied to the specific application context and determines how models perform with regard to the aspect you are interested in.} In this paper, we focus on a generic \textit{real-time} anomaly detection task.

% In general, evaluation criteria employed in existing benchmarks can be roughly categorized into point-based criteria and range-based criteria. \textbf{Point-based} criteria like traditional F1 treat each individual data point as a separate sample, disregarding the holistic characteristics of the anomaly segment. \textbf{Range-based} criteria incorporate segment-level features, for instance, the detection latency~\cite{suite_exa}, into the evaluation. 

In industrial practice, we prefer a criterion that aligns better with the typical workflow of real-time anomaly detection systems deployed in real-world scenarios. In a practical software system, whenever the anomaly score exceeds a predefined threshold, the system triggers an alert to notify the operators. Therefore, an operator tends to prioritize the following key issues: 
\begin{itemize}
\item[\textit{a.}] If a method can always detect anomalies within the anomaly segments, even if the anomaly score surpasses the threshold only once in each segment, it is still considered to possess a significant recall capability.
\item[\textit{b.}] Excessive false alarms would be rather frustrating for operators as they are obligated to examine each alert.
\item [\textit{c.}] The anomaly that lasts longer is likely to be more severe or challenging than the shorter ones.
\item[\textit{d.}] Detecting and addressing anomalies as early as possible can significantly mitigate the economic losses caused by abnormal situations.
\end{itemize}

\begin{figure}[tb]
    \centering
    \includegraphics[width=0.45\textwidth]{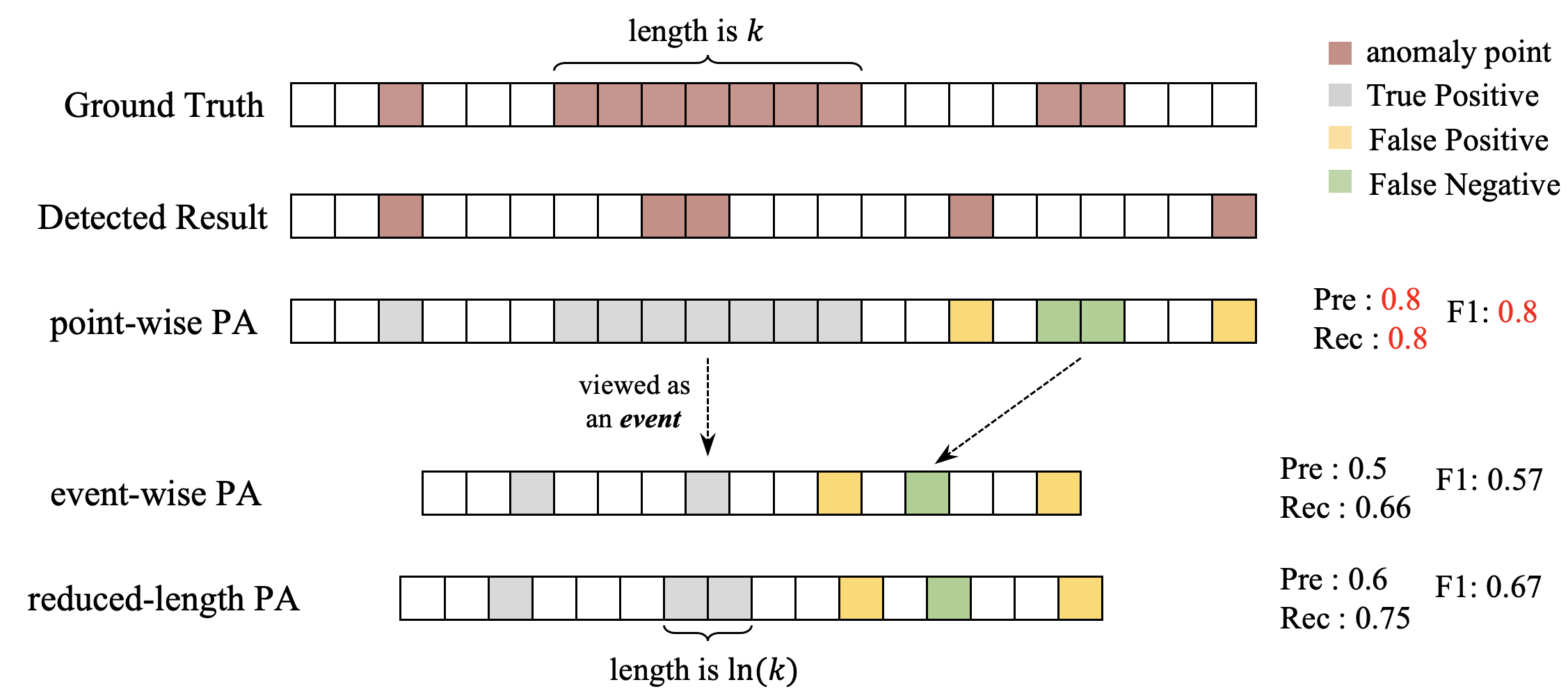}
    \caption{Illustrations of evaluation criteria based on point-adjustment (PA). Point-wise PA gives an inflated score when some anomaly segments persist for a long duration. Event-wise PA treats each anomaly segment as an event, completely disregarding the length of the anomaly segment. Reduced-length PA considers the trade-offs between the two methods, holding greater practical significance in real-world applications.}
    \label{fig.pa}
\end{figure}

Unfortunately, current evaluation criteria~\cite{suite_exa, eval2vus, evaflaw1tisat, evaflaw3aaai, eval1, eval3} fail to simultaneously address all of the above issues. Thus we propose a new criterion that eliminate the effect of the factors that distort the evaluation results by the following strategies:

\noindent\textbf{Point Adjustment.} We utilize a strategy of "adjusting" the output of the algorithm, namely point-adjustment (PA), as the solution to issue \textit{a}. This strategy is widely adopted~\cite{pa1, THOCpa2, omnipa3, usadpa4, interpa5, tfad, cad}. Under this strategy, all timestamps within an anomalous segment are assigned the highest anomaly score present within that segment, thus the whole anomaly segment is considered to be detected if at least one anomaly score surpasses the threshold. Then the F1 score is obtained in a point-based manner (point-wise PA in Fig.~\ref{fig.pa}). 

\noindent\textbf{Anomaly Weights Revision.} However, the naive point-wise PA introduces biased true positives and false negatives. As shown in Fig.~\ref{fig.pa}, as a long anomaly segment is detected under point-adjustment, even if two false alarms are triggered, the Precision is up to 0.8, which contradicts the demand corresponding to issue \textit{b}. The intrinsic flaw of this criteria is that the detection is rewarded generously while pulsing false alarm is penalized just once~\cite{evaflaw1tisat}. Garg et al.~\cite{eventf1} revise the primitive PA from an event perspective. Each anomaly segment is treated as an individual event and contributes to a true positive or false negative only once. This criterion, namely event-wise PA, prevents the inflated evaluation scores originated from illogical criteria, while absolutely overlooking the length of the anomaly segments. Based on the observation \textit{c} , we apply a severity coefficient $\ln{(k+e)}$ to adjust the true positive and false positive measurements associated with the given anomalous segment of length $k$. The criterion is called reduced-length PA, and we categorize it along with event-wise PA as event-based evaluation criteria.

\begin{figure}[htb]
    \centering
    \includegraphics[width=0.45\textwidth]{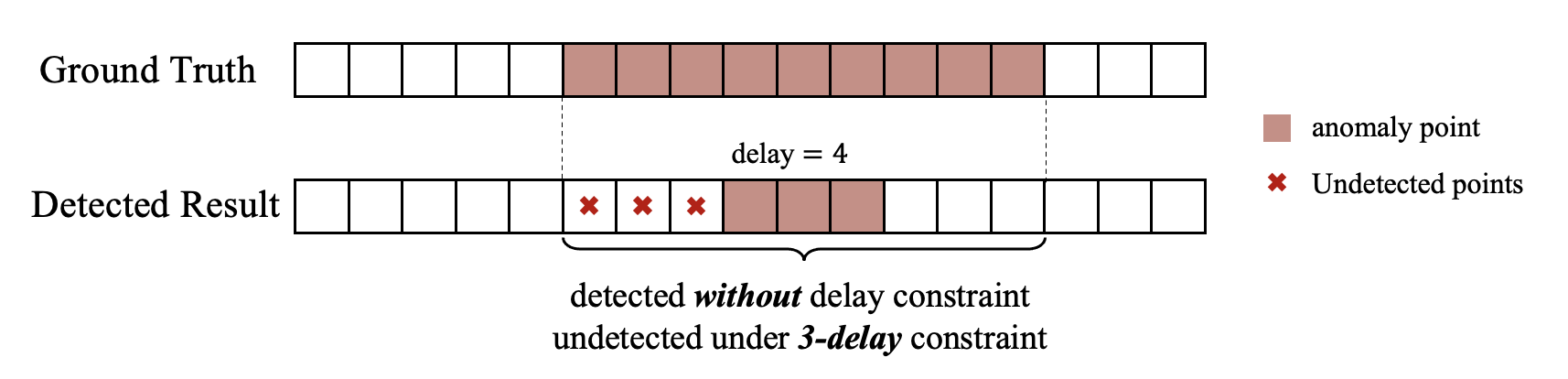}
    \caption{Illustrations of k-delay adjustment. This strategy can be combined with point-adjustment as a complementary evaluation paradigm.}
    \label{fig.kdelay}
\end{figure}

\noindent\textbf{Latency Constraint.} \pch{Issue \textit{d} is addressed by imposing strict constraints on the detection latency. As depicted in the illustration (Fig.~\ref{fig.kdelay}), assuming the latency limit ($k$) is set to 3, an anomaly is considered effectively detected only if it is identified within three sampling points after its occurrence. We designate this strategy as \textit{k-delay adjustment}. This measure enables a more precise assessment of whether the model can meet the requirement of the scenario where there is a high demand for real-time responsiveness. It is equally essential to acknowledge that this approach is applicable only to datasets whose anomalies are labeled without positional bias. Unfortunately, not all datasets can meet this requirement, thus we present the related results as supplementary content.}

\begin{figure}[htb]
    \centering
	\hspace{-5pt}\subfloat[raw segment]{\includegraphics[width=.28\linewidth]{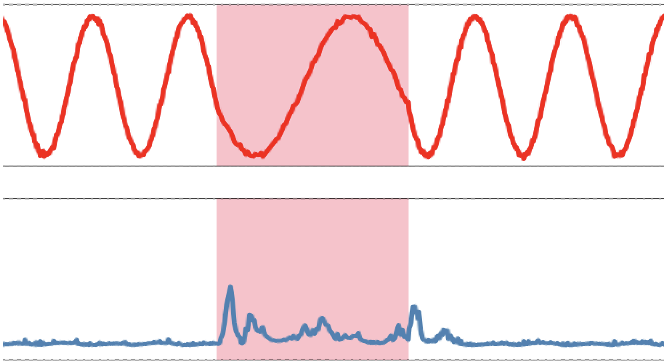}} \hspace{25pt}
    \subfloat[prolonged segment]{\includegraphics[width=.45\linewidth]{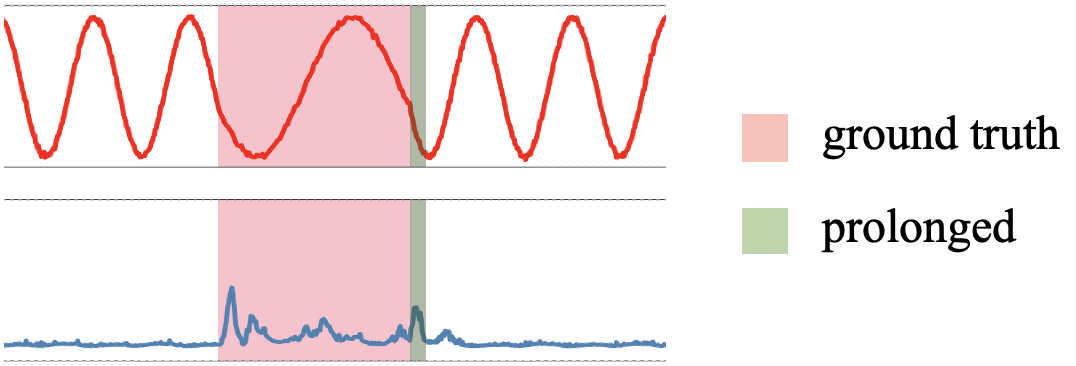}}\hspace{-15pt}
    \caption{Illustrations of prolonging the anomaly segments. The blue line denotes the anomaly scores provided by a particular method. Considering the interference caused by anomalies on the detection of normal values, we should exclude false positives that occur immediately after an anomaly segment.}
    \label{fig.prolong}
\end{figure}

\noindent\textbf{Anomaly Lag Elimination.} Most statistical and deep learning methods generate anomaly scores under prediction-based or reconstruction-based frameworks, both heavily relying on the pattern provided by previous time windows. Hence, the recently concluded anomaly segment, particularly those with longer durations, has the potential to heavily interfere with the subsequent detection process. As shown in Fig.~\ref{fig.prolong}, from a qualitative perspective, we observe that the method accurately detects the frequency anomaly \textit{event}. However, some unexpected false positives emerge due to the aforementioned reasons, resulting in an underestimated evaluation. We slightly extended the anomaly segments (less than 10 time points) to tolerate such occurrences, to make the evaluation more in line with our intuitive comprehension. We carefully handle edge cases to avoid merging two anomaly segments during the operation.

\subsection{Algorithms} 
We adopt 17 semi-supervised/unsupervised methods that have raised significant discussion and wide-ranging impact in the field of univariate time series anomaly detection. 
% The algorithms and implementation details are listed in Appendix~\ref{app:algo}. 
From the perspective of method attributes, these methods can be divided into statistical methods and deep learning methods. 

\noindent\textbf{Statistical methods}. These methods, including Sub-LOF~\cite{sublof}, SAND~\cite{sand}, and MatrixProfile~\cite{MP}, directly compute the minimum similarity between the current and the historical time window. This manner, however, would make the model highly sensitive to the data characteristics in specific scenarios, resulting in a lack of generalizability. 

\noindent\textbf{Deep learning methods}. These methods learn the traits of normal patterns from past data in the training phase and judge whether the current metric is within the normal range in the online detection phase. Furthermore, deep learning methods can be categorized into the following families:

% \begin{itemize}
%     \item \textit{Prediction-based.} These methods aim to predict the "normal" value of the current metric according to the adjacent observations. If the ground truth is far from the predicted value, an anomaly alarm will be triggered.
%     \item \textit{Reconstruction-based.} These methods aim to denoise the anomaly points by the encoding-decoding phase. If the reconstructed value is far from the ground truth, it can be assumed that an anomaly occurs.
%     \item \textit{VAE-based.} These methods believe that the distribution mapping and sampling process can enhance the denoising effect of the model and learn a more robust representation of normal patterns.
%     \item \textit{General Time Series Model.} The designed universal model structure in this class can be used for all time series tasks including time series forecasting, classification, and anomaly detection. They tend to introduce a more general inductive bias that aims to better represent the general features of time series.
% \end{itemize}

\textit{Prediction-based.} These methods aim to predict the "normal" value of the current metric according to the adjacent observations. If the ground truth is far from the predicted value, an anomaly alarm will be triggered. The representative works include AR~\cite{ar}, LSTMAD\footnote{As there is no source code available, we implement LSTMAD in two forms: LSTMAD-$\alpha$ in a seq2seq manner and LSTMAD-$\beta$ in a multi-step prediction manner.}~\cite{lstmad},

\textit{Reconstruction-based.} These methods aim to denoise the anomaly points by the encoding-decoding phase. If the reconstructed value is far from the ground truth, it can be assumed that an anomaly occurs. The representative works include AE~\cite{ae}, EncDecAD~\cite{encdec}, SRCNN~\cite{srcnn}, Anomaly Transformer~\cite{anotrans}, TFAD~\cite{tfad}, and TranAD~\cite{tranad}.

\textit{VAE-based.} These methods hold the assumption that the distribution mapping and sampling process can enhance the denoising effect of the model and learn a more robust representation of normal patterns. The representative works include Donut~\cite{donut} and FCVAE~\cite{fcvae}.

\textit{General Time Series Model.} The designed universal model structure in this class can be used for all time series tasks including time series forecasting, classification, and anomaly detection. They tend to introduce a more general inductive bias that aims to better represent the general features of time series. The representative works include TimesNet~\cite{timesnet}, OFA~\cite{ofa} and FITS~\cite{fits}.

\section{Experimental settings}
\label{app:setting}

\subsection{Experimental Platform} All experiments are performed on a server equipped with Dual Intel(R) Xeon(R) Silver 4316 (12-core) and 256GB RAM. The operating system of this server is Ubuntu 22.04 LTS. An NVIDIA GeForce RTX 3090 24GB GDDR6 GPU is utilized to accelerate the training and inference processes of all models.

\subsection{Datasets} With the exception of the UCR dataset and AIOPS dataset where the training and testing sets are already specified, we partition each time series into training, validation, and test sets following a 4:1:5 ratio. Given that anomalies in some datasets are randomly distributed, the test set of some sequences does not contain any anomalies after partitioning. Therefore, we exclude these anomaly-free sequences from consideration. Due to the fact that the original implementation of TODS sometimes generates anomalies that do not match their specified anomaly types (e.g., constructing trend and seasonal anomalies may result in obvious global outliers), we modify the anomaly generation code in TODS to ensure that the injected anomalies align more closely with their defined types. It is worth noting that we aggregate the overall evaluation at the dataset level instead of the curve level because the imbalance in sample sizes across different datasets can lead to biased results. 

\subsection{Learning Schema} We evaluated existing methods under all proposed learning schemas to obtain a more comprehensive understanding of the model's performance from different perspectives. Statistical methods are not assessed under all-in-one and zero-shot schema due to assumption conflict. Under the all-in-one schema, taking the UCR dataset as an example, we mix all samples together from all time series' training sets during the training phase. Under zero-shot schema, we use a fixed random seed to split UCR into two subsets, each of which includes half of the time series. We mix all training samples from one subset, and the other acts as the test set. All methods share the same training set in zero-shot mode.

\subsection{Evaluation Metrics} Following the event-based evaluation criterion outlined in Sec.~\ref{sec:eva_cri}, we establish two concrete metrics, namely $F1_{best}$ and $AUPRC$. $F1_{best}$ denotes the highest F1 score calculated under reduced-length point-adjustment when iterating over all possible thresholds. $AUPRC$ calculates the area under the precision-recall curve generated according to reduced-length point-adjustment, which is widely applied in cases where class imbalance is present in the data. While $F1_{best}$ can measure the best detection performance that the model can achieve on the current test set, $AUPRC$ provides a more nuanced evaluation of the model's performance across different levels of recall, which can be important in anomaly detection. A model with a higher $AUPRC$ tends to be more robust.

\subsection{Implementation details of Algorithms}

All methods are implemented in Sklearn or Pytorch, either based on open-source repositories\footnote{SRCNN comes from \url{https://github.com/microsoft/anomalydetector}, AnomalyTransformer comes from \url{https://github.com/thuml/Anomaly-Transformer}, TimesNet comes from  \url{https://github.com/thuml/Time-Series-Library/blob/main/models/TimesNet.py}, Donut comes from \url{https://github.com/wagner-d/TimeSeAD/blob/master/timesead/models/generative/donut.py}, TFAD comes from \url{https://github.com/DAMO-DI-ML/CIKM22-TFAD}, OFA comes from \url{https://github.com/DAMO-DI-ML/NeurIPS2023-One-Fits-All}, FITS comes from \url{https://github.com/VEWOXIC/FITS}, } or reproduced based on the original paper's description. All methods are integrated into the TimeSeriesBench suite. If the model has specific hyperparameters set for a particular dataset, we use the parameters specified for that dataset. Otherwise, we use the default hyperparameters provided in the source code. The early stopping mechanism is applied to all methods on the validation set.

\newcounter{RI}

\section{Experimental Results}
\label{sec:result}
In this section we conduct a rigorous analysis of the model's effectiveness under various settings, aiming to provide meaningful research insights (\textbf{RI})\stepcounter{RI} from unprecedented perspectives that can have implications for the design and application of methods. We aim to provide insights into the following research questions: 
\begin{itemize}
    \item[\textit{1.}] How is the overall performance of the models, and what factors contribute to these results?
    \item[\textit{2.}] How does the model's performance vary under different learning schemas?
    \item[\textit{3.}] How does the model's performance vary in detecting different types of anomalies?
\end{itemize}
\subsection{Overall Performance}
\begin{figure*}[tb]
    \centering
    \hspace{20pt}
	\subfloat[naive]{\includegraphics[width=.36\textwidth]{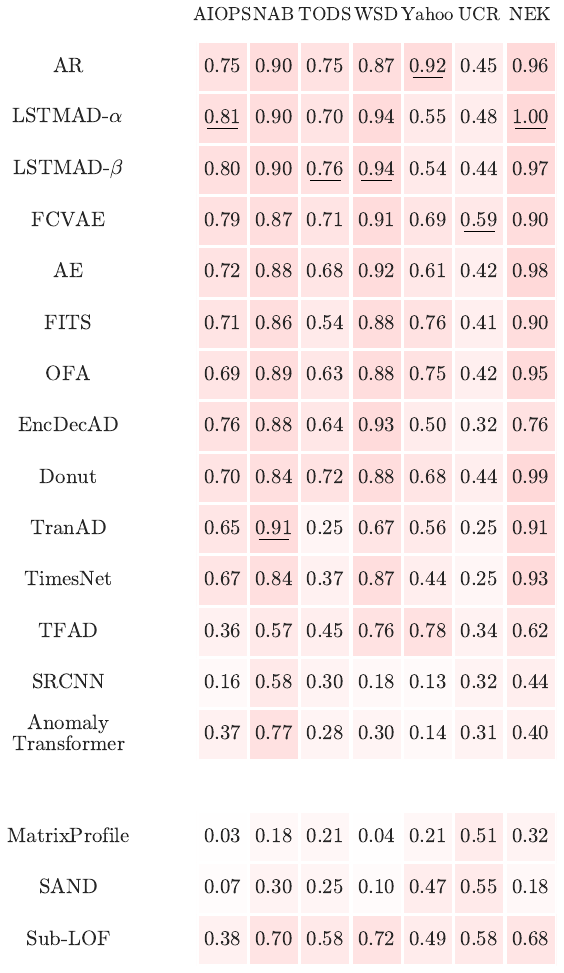}} 
    \hfill
    \subfloat[all-in-one]{\includegraphics[width=.241\textwidth]{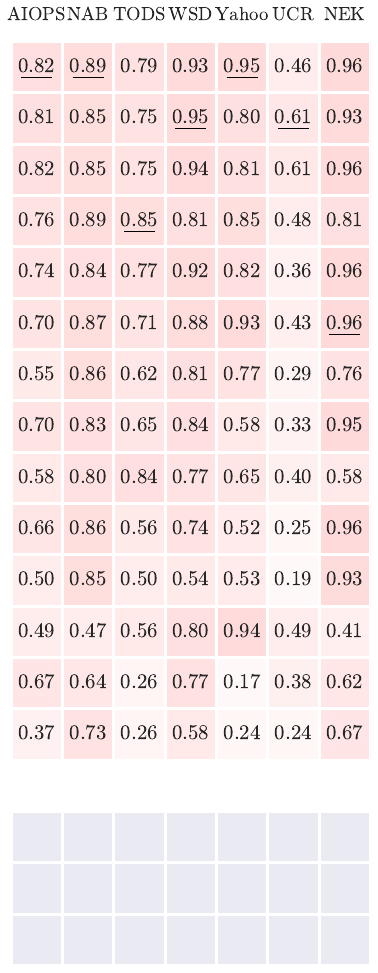}}
    \hfill 
    \subfloat[zero-shot]{\includegraphics[width=.330\textwidth]{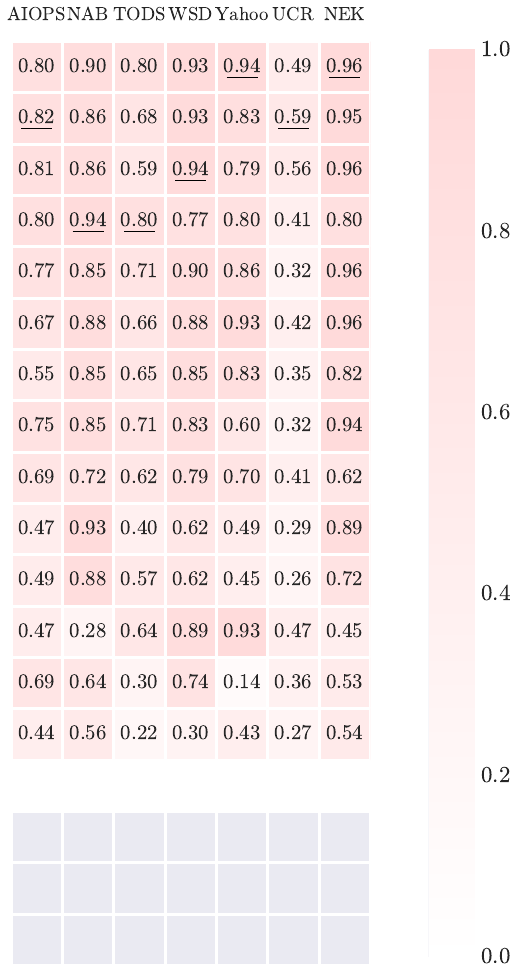}}
    \hspace{20pt}
    
    \caption{Overall performance ranking using different learning schema under reduced-length $F1_{best}$. Methods are ranked in descending order according to the average of all scores. The row names denote the names of methods, while the column names denote the names of the datasets. The best score of each column is underlined. Due to the inherent assumptions behind the statistical methods being incompatible with the all-in-one and zero-shot schemas, these methods are not included in the ranking, and the related score cells are set to grey without values.}
    \label{fig.overallp}
\end{figure*}

The overall ranking list is presented in Fig.~\ref{fig.overallp}, including the performance of each method using naive, all-in-one, and zero-shot schema, respectively. Since statistical methods, especially those based on streaming, heavily rely on the premise of data being identically distributed, these methods are not well-suited for the all-in-one and zero-shot learning schemas. Therefore, we only evaluate them under the naive schema. First and foremost, the performance of various models under the naive and all-in-one learning schema can exhibit substantial disparities \textbf{(RI \the\value{RI})}\stepcounter{RI}. \pch{For example, on the NEK dataset, although Donut performs well on the naive schema, its performance significantly degrades under the all-in-one schema. In contrast, FITS achieve better performance under all-in-one schema on this dataset. We list the impacts of different learning schemas on the performance of different methods in Table~\ref{tb:comp}.} \textit{It is thought-provoking that these differences do not show a clear bias, suggesting that these variations are the result of a combination of multifaceted factors, which is discussed in detail in Sec.~\ref{sec:finegrain}. }Additionally, given the consideration of errors arising from random dataset partitioning, the models demonstrated a relatively consistent performance across the all-in-one and zero-shot modes, as there may exist interdependence among time series in the same dataset \textbf{(RI \the\value{RI})}\stepcounter{RI}.

\begin{table*}[htb]
    \centering
    \caption{Performance differences on all datasets using different learning schemas. Best scores are highlighted in bold, and second-best scores are highlighted in bold and underlined. Performance improvements exceeding 5\% are marked with {\color{blue} $\uparrow$}, while performance declines exceeding 5\% are marked with {\color{red} $\downarrow$}.}
    \resizebox{\linewidth}{!}{
    \begin{tabular}{c|cc|cc|cc|cc|cc|cc|cc}
    \hline
        \multirow{2}{*}{\textbf{Method}} & \multicolumn{2}{c|}{AIOPS} & \multicolumn{2}{c|}{NAB} & \multicolumn{2}{c|}{TODS} & \multicolumn{2}{c|}{WSD} & \multicolumn{2}{c|}{Yahoo} & \multicolumn{2}{c|}{UCR} & \multicolumn{2}{c}{NEK} \\
         & $S_{one}$ & $S_{all}$ & $S_{one}$ & $S_{all}$ & $S_{one}$ & $S_{all}$ & $S_{one}$ & $S_{all}$ & $S_{one}$ & $S_{all}$ & $S_{one}$ & $S_{all}$ & $S_{one}$ & $S_{all}$ \\
    \hline
        AR & 0.7517 & \textbf{0.8214}{\color{blue} $\uparrow$} & \textbf{\underline{0.9035}} & \textbf{0.8881} & \textbf{\underline{0.7472}} & 0.7859{\color{blue} $\uparrow$} & 0.868 & 0.9276{\color{blue} $\uparrow$} & \textbf{0.9195} & \textbf{0.948} & 0.4454 & 0.4636 & 0.9555 & \textbf{\underline{0.9612}} \\
        LSTMAD-$\alpha$ & \textbf{0.8081} & 0.8141 & 0.8984 & 0.8492{\color{red} $\downarrow$} & 0.7015 & 0.7538{\color{blue} $\uparrow$} & \textbf{\underline{0.9381}} & \textbf{0.9467} & 0.5496 & 0.8{\color{blue} $\uparrow$} & \textbf{\underline{0.4836}} & \textbf{0.6144}{\color{blue} $\uparrow$} & \textbf{0.9971} & 0.9349{\color{red} $\downarrow$} \\
        LSTMAD-$\beta$ & \textbf{\underline{0.8034}} & \textbf{\underline{0.8151}} & 0.9031 & 0.8484{\color{red} $\downarrow$} & \textbf{0.7589} & 0.7495 & \textbf{0.939} & \textbf{\underline{0.9445}} & 0.5366 & 0.8114{\color{blue} $\uparrow$} & 0.4366 & \textbf{\underline{0.6101}}{\color{blue} $\uparrow$} & 0.9744 & 0.961 \\
        AE & 0.7222 & 0.7448 & 0.8751 & 0.8375 & 0.6842 & 0.7712{\color{blue} $\uparrow$} & 0.918 & 0.9192 & 0.612 & 0.8205{\color{blue} $\uparrow$} & 0.4198 & 0.3644{\color{red} $\downarrow$} & 0.9801 & 0.9609 \\
        EncDecAD & 0.7583 & 0.7003{\color{red} $\downarrow$} & 0.8752 & 0.8336 & 0.6359 & 0.6534 & 0.9285 & 0.8353{\color{red} $\downarrow$} & 0.5018 & 0.5842{\color{blue} $\uparrow$} & 0.3178 & 0.3333 & 0.755 & 0.9534{\color{blue} $\uparrow$} \\
        Donut & 0.6957 & 0.5827{\color{red} $\downarrow$} & 0.8397 & 0.7966{\color{red} $\downarrow$} & 0.7207 & \textbf{\underline{0.836}}{\color{blue} $\uparrow$} & 0.8787 & 0.7666{\color{red} $\downarrow$} & 0.6813 & 0.6498 & 0.4409 & 0.4{\color{red} $\downarrow$} & \textbf{\underline{0.9885}} & 0.5801{\color{red} $\downarrow$} \\
        FCVAE & 0.7851 & 0.7593 & 0.874 & \textbf{\underline{0.8857}} & 0.7145 & \textbf{0.8526}{\color{blue} $\uparrow$} & 0.9087 & 0.8121{\color{red} $\downarrow$} & 0.6883 & 0.8537{\color{blue} $\uparrow$} & \textbf{0.5873} & 0.4766{\color{red} $\downarrow$} & 0.8977 & 0.8148{\color{red} $\downarrow$} \\
        SRCNN & 0.1627 & 0.6672{\color{blue} $\uparrow$} & 0.5802 & 0.635{\color{blue} $\uparrow$} & 0.3042 & 0.2598{\color{red} $\downarrow$} & 0.1785 & 0.7742{\color{blue} $\uparrow$} & 0.1261 & 0.1728{\color{blue} $\uparrow$} & 0.3227 & 0.3791{\color{blue} $\uparrow$} & 0.4365 & 0.6173{\color{blue} $\uparrow$} \\
        AnomalyTransform & 0.3728 & 0.3656 & 0.7739 & 0.7322{\color{red} $\downarrow$} & 0.2827 & 0.2631{\color{red} $\downarrow$} & 0.2984 & 0.5838{\color{blue} $\uparrow$} & 0.145 & 0.2377{\color{blue} $\uparrow$} & 0.3104 & 0.2372{\color{red} $\downarrow$} & 0.3997 & 0.666{\color{blue} $\uparrow$} \\
        TFAD & 0.3551 & 0.4889{\color{blue} $\uparrow$} & 0.5746 & 0.4679{\color{red} $\downarrow$} & 0.4526 & 0.5626{\color{blue} $\uparrow$} & 0.7561 & 0.8016{\color{blue} $\uparrow$} & \textbf{\underline{0.781}} & \textbf{\underline{0.9361}}{\color{blue} $\uparrow$} & 0.3398 & 0.4941{\color{blue} $\uparrow$} & 0.625 & 0.4103{\color{red} $\downarrow$} \\
        TranAD & 0.6486 & 0.6561 & \textbf{0.9101} & 0.8567{\color{red} $\downarrow$} & 0.2531 & 0.5624{\color{blue} $\uparrow$} & 0.6738 & 0.7409{\color{blue} $\uparrow$} & 0.562 & 0.5211{\color{red} $\downarrow$} & 0.2527 & 0.2483 & 0.9071 & 0.9571{\color{blue} $\uparrow$} \\
        TimesNet & 0.6737 & 0.4988{\color{red} $\downarrow$} & 0.8419 & 0.8544 & 0.3695 & 0.5032{\color{blue} $\uparrow$} & 0.8684 & 0.539{\color{red} $\downarrow$} & 0.4444 & 0.5341{\color{blue} $\uparrow$} & 0.2493 & 0.1856{\color{red} $\downarrow$} & 0.9335 & 0.9257 \\
        OFA & 0.6891 & 0.5544{\color{red} $\downarrow$} & 0.8869 & 0.861 & 0.6263 & 0.6187 & 0.8784 & 0.8111{\color{red} $\downarrow$} & 0.7512 & 0.766 & 0.4169 & 0.2888{\color{red} $\downarrow$} & 0.9471 & 0.7642{\color{red} $\downarrow$} \\
        FITS & 0.7139 & 0.6986 & 0.8617 & 0.8683 & 0.5371 & 0.7146{\color{blue} $\uparrow$} & 0.8796 & 0.8788 & 0.7564 & 0.9261{\color{blue} $\uparrow$} & 0.4092 & 0.428 & 0.8975 & \textbf{0.9625}{\color{blue} $\uparrow$} \\

    \hline
    \end{tabular}
    }
    \label{tb:comp}
\end{table*}
\begin{figure}[htb]
    \centering
    \includegraphics[width=0.48\textwidth]{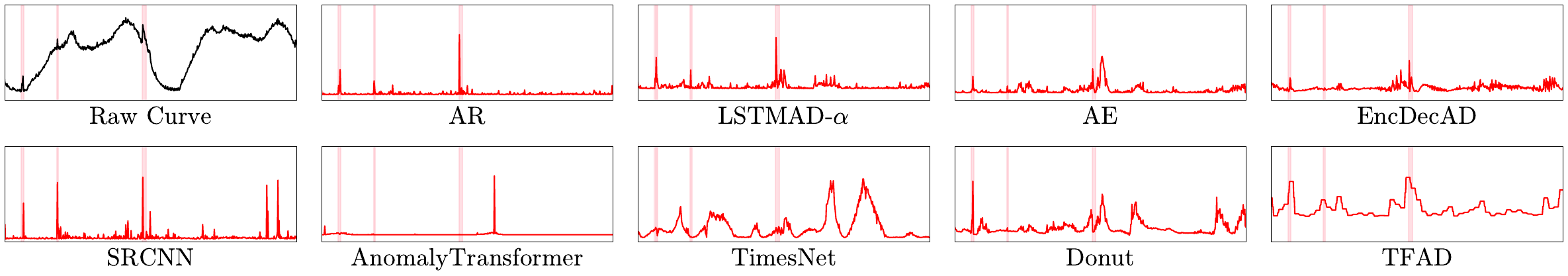}
    \caption{Illustrations that some methods fail to effectively detect point-wise anomalies in certain scenarios. The black line represents the original curve, while the red line represents the anomaly scores provided by the method. Anomalous segments are highlighted in pink.}
    \label{fig.kpi1}
\end{figure}

Generally, the statistical methods exhibit great performance on the dataset with relatively stable shapelets and low noise level (UCR), but work poor on other datasets with more noise. For deep learning based models, contrary to our expectations, the overall performances of the models that contain more complicated structures are not satisfactory \textbf{(RI \the\value{RI})}\stepcounter{RI}. 
% \del{According to our examination across existing datasets, a substantial portion of anomalies in real-world datasets exhibit noticeable deviations from their very close neighboring points (sudden spikes or drops), especially for AIOPS, WSD, and Yahoo. These anomalies can be categorized as either contextual outliers or global outliers, both of them belonging to point-wise anomaly type. Due to its ability to capture very local temporal correlations and its ease of convergence, AR surpasses a number of modern networks in cases where point-wise anomalies dominate. }
We present a case (Fig.~\ref{fig.kpi1}) from the AIOPS dataset to vividly demonstrate this phenomenon. Even though these point-wise anomalies seem to be quite evident, a significant portion of the methods are unable to handle such events effectively. As numerous factors can lead to a decrease in performance for DNNs, we summarize the major factors that result in the underperformance of some deep learning methods in anomaly detection tasks: 

\pch{\noindent\textbf{Poor noise resistance.} The models' anti-noise ability plays an important role in AD tasks. Compared to other domains like NLP or CV, time series data is affected by more factors during the collection process, resulting in more and intractable noise. Although models using simple structures (AR, FITS) sacrifices some feature expression capability, it's hard for them to overfit noise. This contributes to the good results. For LSTM-series methods, due to the iterative encoding process, the forget gate in LSTMs is more likely to minimize the retention of low signal-to-noise ratio data in the hidden state. In constrast, due to the lack of special design for anti-noise, simply employing complicated backbones like transformer may lead to the overfitting of the noise, and finally result in the poor performance as shown in Fig.~\ref{fig.kpi1} \textbf{(RI \the\value{RI})}\stepcounter{RI}.}

\begin{figure}[htbp]
    \centering
    \subfloat[Lacking data under one-by-one schema]{
    \includegraphics[width=0.45\textwidth]{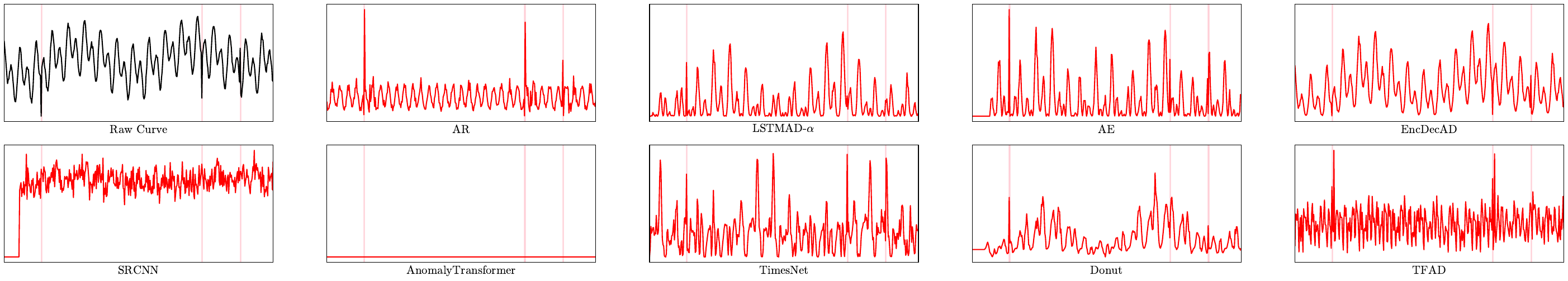}
    }
    \vspace{5mm}
    \\
    \subfloat[Enriching data by using all-in-one schema]{
    \includegraphics[width=0.45\textwidth]{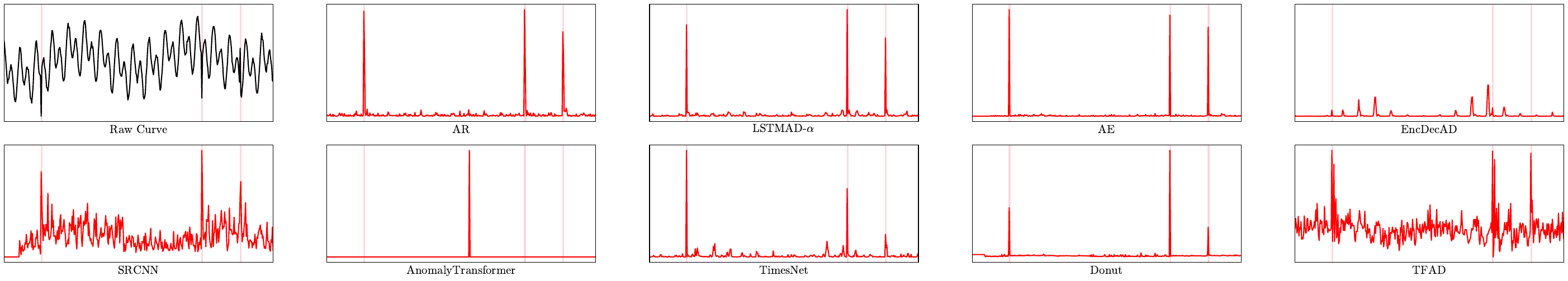}
    }
    \caption{Case study on Yahoo\_A3Benchmark-TS13 shows that lack of training data induces poor performance. }
    \label{fig.lack}
\end{figure}

\noindent\textbf{Lack of training data.}\pch{ As shown in Fig.~\ref{fig.lack}, some models, especially the ones with complicated structures, achieve significant improvement in performance when using more training data. Models with complicated structures usually require a larger volume of training data to avoid overfitting due to their higher degrees of freedom and flexibility \textbf{(RI \the\value{RI})}\stepcounter{RI}.} Similarly, models that employ intricate loss functions often necessitate a greater amount of training data, including diverse examples, to prevent the emergence of falling into local minima and even trivial solutions.

\noindent\textbf{Trends with large periods.} Some data may exhibit long-term trends or patterns that span large periods. These trend components can cause the data distribution to deviate from the comfort zone of the model, causing the model to calculate anomaly scores based on a biased data distribution it has never encountered before \textbf{(RI \the\value{RI})}\stepcounter{RI}. 

\begin{figure}[htb]
    \centering
    \includegraphics[width=0.45\textwidth]{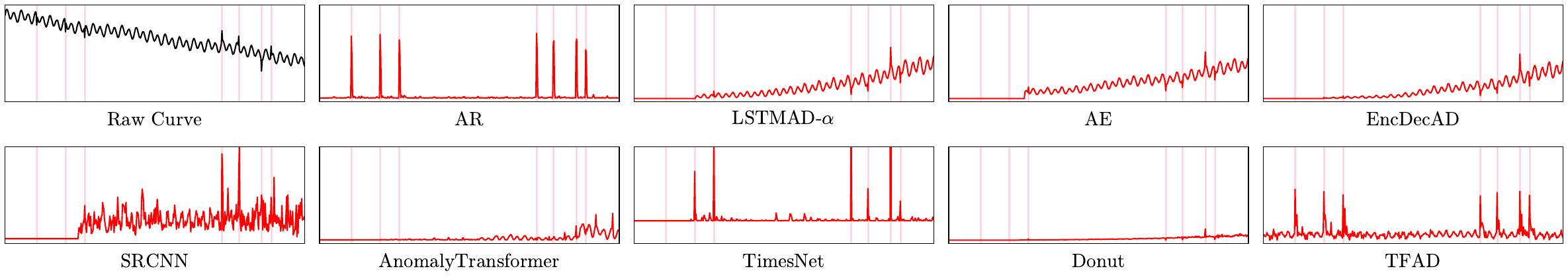}
    \caption{Case study on Yahoo\_A3Benchmark-TS14 shows that trends with large periods induce poor performance. The scores are obtained under the all-in-one learning schema, eliminating the impact of insufficient data.}
    \label{fig.trend}
\end{figure}

\noindent\textbf{Unreasonable inductive bias.} Deep learning models rely on their architectural design and optimization algorithms to detect anomalies in a semi-supervised manner. If the chosen model architecture or optimization approach is not well-suited for point-wise anomaly detection, the models may have an unreasonable inductive bias that hampers their ability to accurately detect anomalies \textbf{(RI \the\value{RI})}\stepcounter{RI}. 

\pch{Based on the above observations, it is imperative for newly proposed methods to place greater emphasis on these factors to avoid poor results on these distinct anomalies. We believe that the potential of the powerful representation capability of complex structures has not been fully explored yet in time series anomaly detection tasks. We strongly advocate prioritizing the refinement of module design or loss function in order to enhance the models' resilience to noise and their capacity for generalization. Thus complicated backbones can better leverage their expressive power for temporal features without overfitting to noise.}

\subsection{Performance in Fine-Grained Scenarios}
\label{sec:finegrain}

\begin{figure}[htbp]
    \centering
    \subfloat[naive \& point-wise]{
    \includegraphics[width=0.45\linewidth]{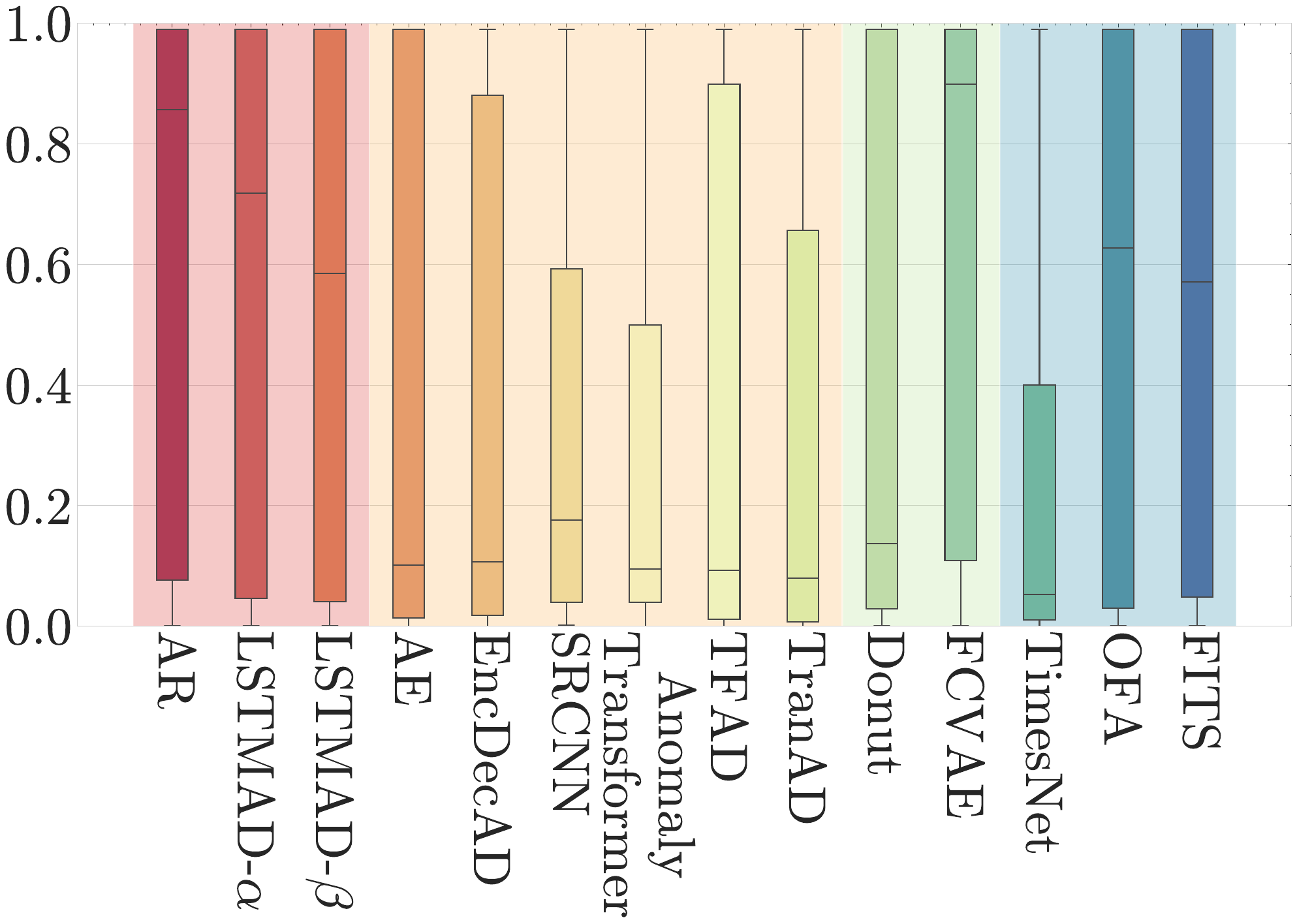}
    } \hspace{5pt}
    \subfloat[naive \& pattern-wise]{
    \includegraphics[width=0.45\linewidth]{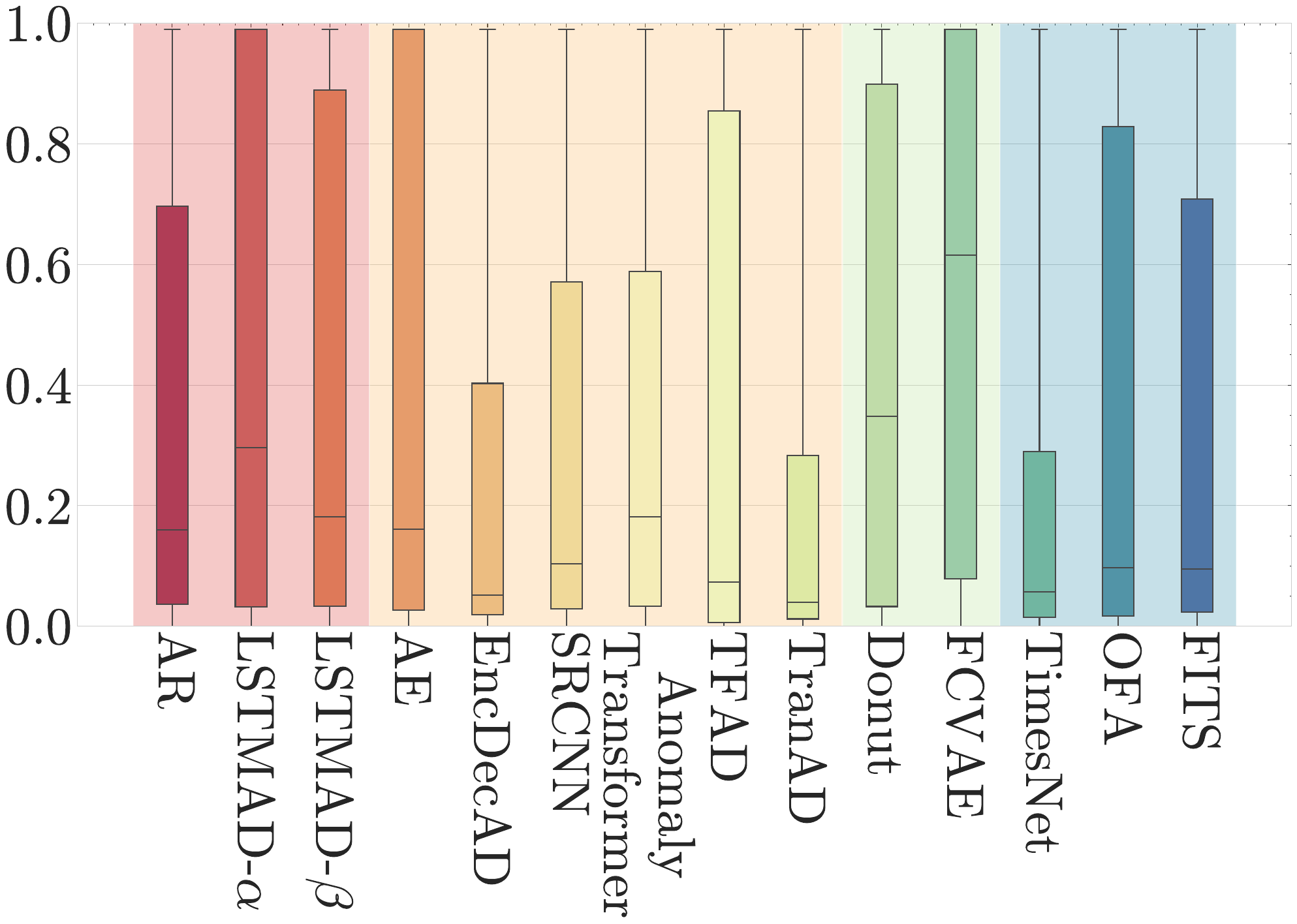}
    } 
    \vspace{3mm}
    \\
    \subfloat[all-in-one \& point-wise]{
    \includegraphics[width=0.45\linewidth]{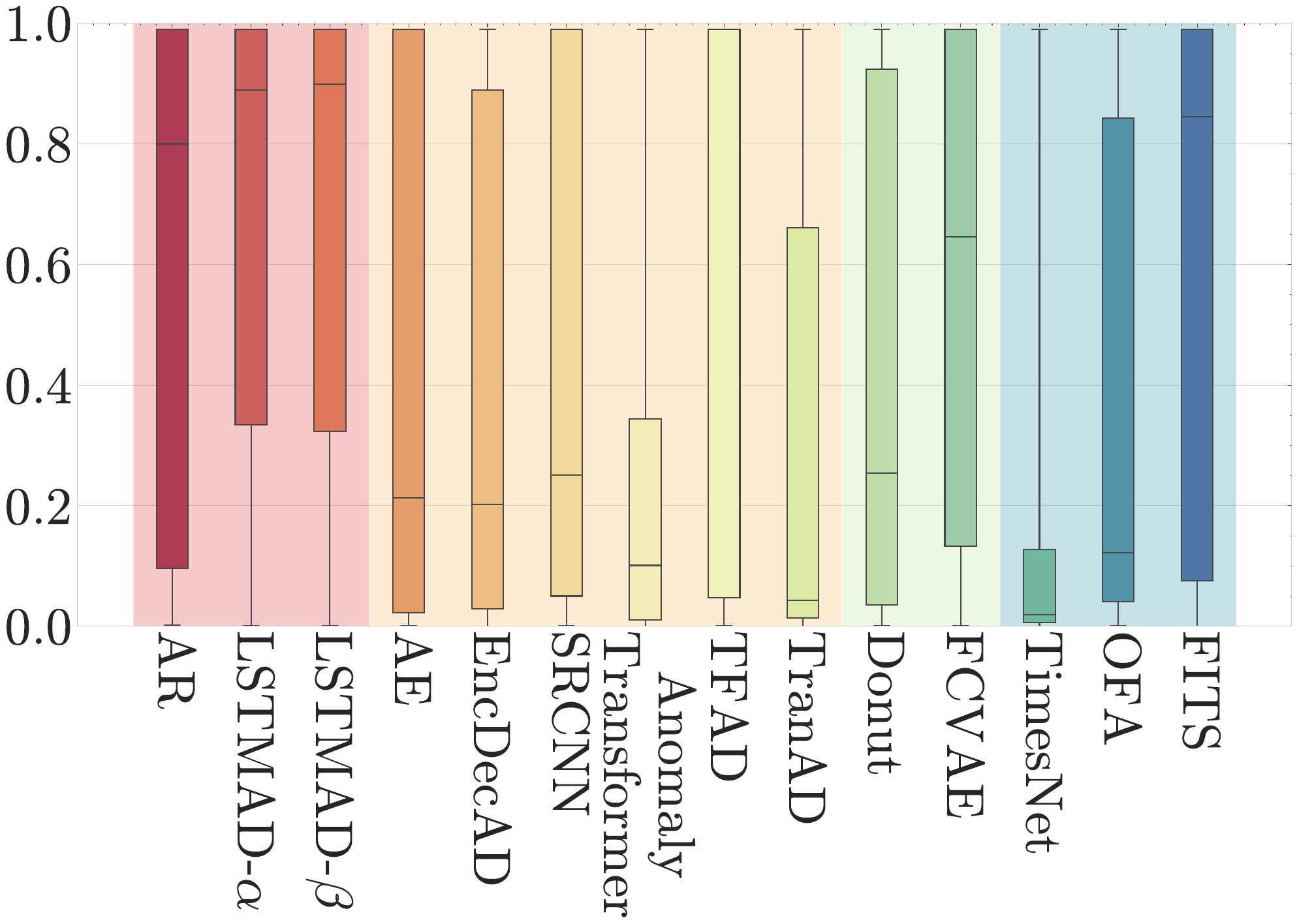}
    } \hspace{5pt}
    \subfloat[all-in-one \& pattern-wise]{
    \includegraphics[width=0.45\linewidth]{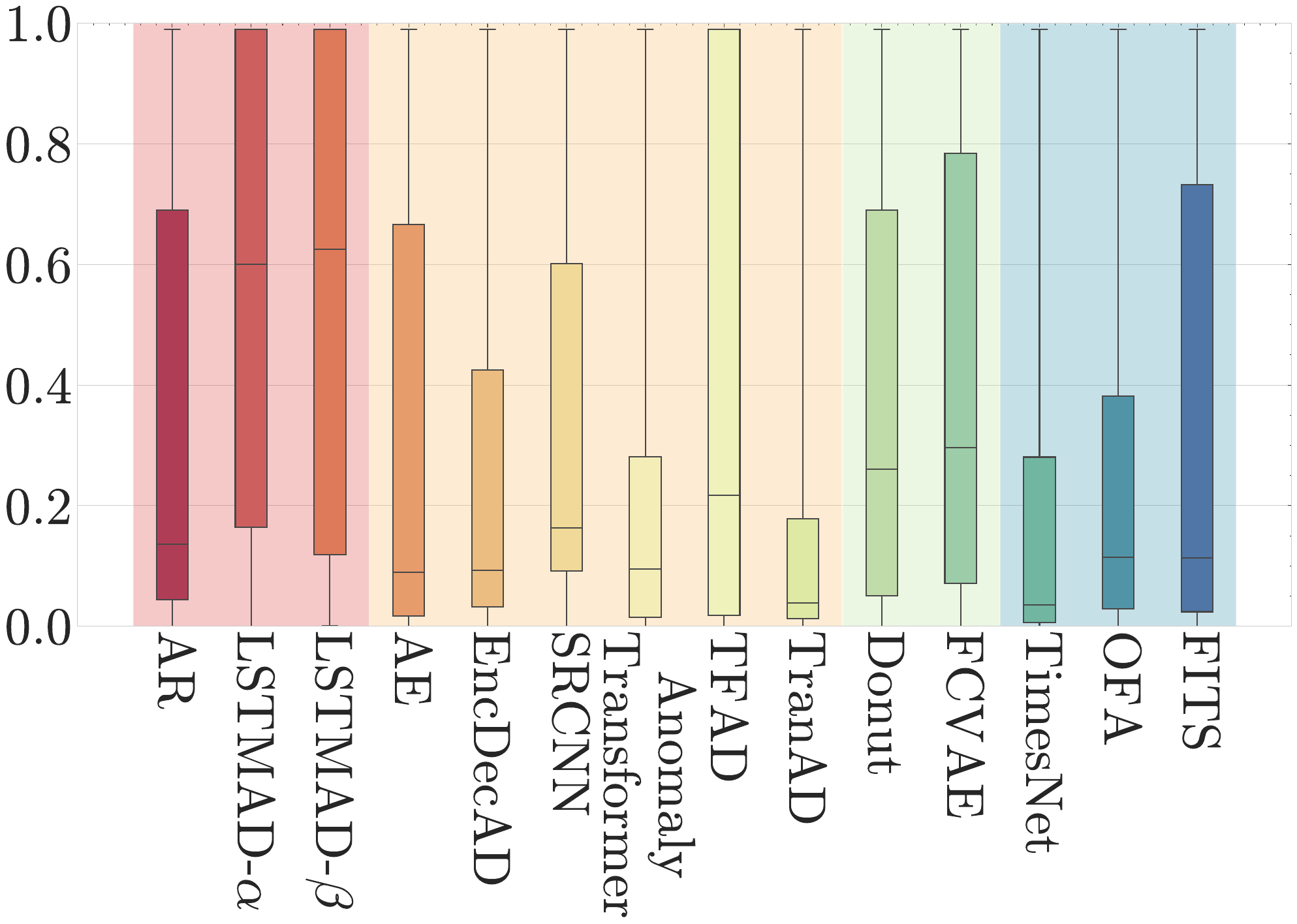}
    } 
    % \vspace{5mm}
    \caption{The 2x2 matrix reflecting the variations in model performance under different conditions. We denote prediction, reconstruction, VAE-based, and general time series methods in light red, yellow, green, and blue backgrounds.}
    \label{2x2}
\end{figure}

Merely referring to aggregated data at the granularity of datasets is insufficient to gain a profound understanding of the model's performance. We carry out a fine-grained comparison of model performance from the perspectives of both learning schema and anomaly type. The UCR dataset is widely acknowledged for its comprehensive annotation of various anomaly types, making it a well-labeled dataset. Additionally, each time series in the UCR dataset is guaranteed to contain only one anomaly, which facilitates the categorization of different anomaly types. To further refine our analysis, we partition the UCR dataset into two subsets based on its supplemental materials~\cite{UCR_sm}: one subset exclusively consisting of point-wise anomalies and another subset focusing solely on pattern-wise anomalies. We form a 2x2 matrix to reflect the variations in model performance under different conditions by incorporating the dimensions of "naive" and "all-in-one" learning schemas, along with different types of anomalies, as shown in Fig.~\ref{2x2}. 

\begin{figure}[tbp]
    \centering
    \subfloat[Model comparison under naive learning schema]{
    \includegraphics[width=0.45\textwidth]{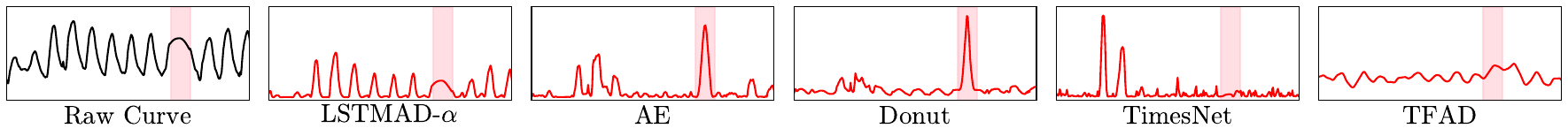}
    } \vspace{3mm}
    \\
    \subfloat[Model comparison under all-in-one learning schema]{
    \includegraphics[width=0.45\textwidth]{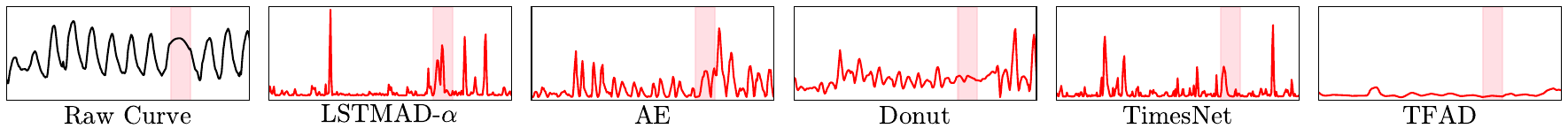}
    } 
    % \vspace{5mm}
    \caption{Case study on CIMIS44AirTemperature2 which contains an easily understandable pattern-wise anomaly.}
    \label{fig.tmp}
\end{figure}

From the boxplots, no solution can always outperform the others in all situations \textbf{(RI \the\value{RI})}\stepcounter{RI}. It is also apparent that the majority of models exhibit a significantly higher detection performance for point-wise anomalies compared to pattern-wise anomalies regardless of the learning schema employed \textbf{(RI \the\value{RI})}\stepcounter{RI}. On a more detailed level, prediction-based models excel in identifying point-wise anomalies, as the steep peak/valley is unpredictable in most cases \textbf{(RI \the\value{RI})}\stepcounter{RI}. Due to the inclusion of a more diverse data distribution under all-in-one learning schema, the models are more likely to learn robust representations of patterns. As a result, under the all-in-one learning schema, there is a significant improvement in the detection performance of point-wise anomalies \textbf{(RI \the\value{RI})}\stepcounter{RI}. This observation is particularly pronounced in the case of the Yahoo dataset each time series of which is relatively short. With regard to pattern-wise anomalies which are considered to be more challenging, the situation becomes more complex and interesting. Methods that aim to generate the temporal window after projecting them to low-dimensional representations (Donut and FCVAE) surpass all others in detecting pattern-wise anomalies, while the performance greatly declines when under the all-in-one mode  \textbf{(RI \the\value{RI})}\stepcounter{RI}. The former indicates that the assumption of low-dimensional representations hardly reconstructing high-level pattern-wise anomaly indeed works. An illustration is presented to confirm this in Fig.~\ref{fig.tmp}(a), where Donut can easily handle the "smooth" high-level anomaly. However, the performance deteriorates or even becomes ineffective under the all-in-one mode (Fig.~\ref{fig.tmp}(b)). We hypothesize that the capacity of low-rank representations may struggle to cover the diverse data distributions present in various curves. Also, although the general time series methods (TimesNet, OFA, FITS) perform well in other time series tasks, they struggle to outperform methods specifically designed for anomaly detection tasks like LSTMAD and FCVAE  \textbf{(RI \the\value{RI})}\stepcounter{RI}. This may indicate that there is a gap between time series anomaly detection tasks and general time representation tasks. For example, anomaly detection tasks may require models to have a stronger denoising effect. These gaps need special attention when designing models for anomaly detection tasks.

\subsection{Performance under Delay-Constraint}
Issue \textit{d} in Sec.~\ref{sec:eva_cri} demands exceptionally high data set quality, whereas some datasets (including NAB, UCR, Yahoo) fail to offer labels without positional bias. Therefore, the result merely provide a approximate reference on these datasets.
We manually set a relatively suitable latency limit ($K$) for each dataset based on the sampling frequency and label quality of the dataset to evaluate the performance of each method under strict real-time requirements. The overall results are shown in Table~\ref{tb:delay}. Compared to the results without delay-constraint, it can be seen that the performance of some methods significantly deteriorates in this setup, even approaching a 50\% decrease (e.g., Anomaly Transformer on TODS). Although EncDecAD is inferior to FITS under all-in-one schema in Fig.~\ref{fig.overallp}, when considering delay-constraint, EncDecAD outperforms FITS on TODS, suggesting that some methods may have greater potential in terms of early detection capabilities.

\begin{table*}[htb]
    \centering
    \caption{Performance comparison using all-in-one learning schema under K-delay reduced-length $F1_{best}$ and $AUPRC$}
    \resizebox{\linewidth}{!}{
    \begin{tabular}{c|cc|cc|cc|cc|cc|cc|cc}
    \hline
        \multirow{2}{*}{\textbf{Method}} & 
        \multicolumn{2}{c|}{AIOPS (K=10)} & \multicolumn{2}{c|}{NAB (K=150)} & \multicolumn{2}{c|}{TODS (K=3)} & \multicolumn{2}{c|}{WSD (K=10)} & \multicolumn{2}{c|}{Yahoo (K=3)} & \multicolumn{2}{c|}{UCR (K=50)} & \multicolumn{2}{c}{NEK (K=10)} \\
        & $F1_{best}$ & $AUPRC$ & $F1_{best}$ & $AUPRC$ & $F1_{best}$ & $AUPRC$ & $F1_{best}$ & $AUPRC$ & $F1_{best}$ & $AUPRC$ & $F1_{best}$ & $AUPRC$ & $F1_{best}$ & $AUPRC$ \\
    \hline
        MatrixProfile & - & - & - & - & - & - & - & - & - & - & - & - & - & -  \\
        SAND & - & - & - & - & - & - & - & - & - & - & - & - & - & -  \\
        Sub-LOF & - & - & - & - & - & - & - & - & - & - & - & - & - & -  \\
        AR & 0.7815 & 0.766 & \textbf{\underline{0.5334}} & \textbf{\underline{0.407}} & \textbf{0.6748} & \textbf{0.6827} & 0.7251 & 0.6888 & \textbf{0.9451} & \textbf{0.9345} & 0.3422 & 0.3121 & 0.9137 & 0.9154 \\
        LSTMAD-$\alpha$ & \textbf{\underline{0.7819}} & \textbf{\underline{0.7924}} & 0.4235 & 0.2928 & \textbf{\underline{0.6437}} & \textbf{\underline{0.6424}} & \textbf{0.7627} & \textbf{0.7276} & 0.7946 & 0.7767 & \textbf{0.4494} & \textbf{\underline{0.4063}} & 0.9196 & 0.9308 \\
        LSTMAD-$\beta$ & \textbf{0.7826} & \textbf{0.7957} & 0.4275 & 0.2928 & 0.6234 & 0.63 & \textbf{\underline{0.7593}} & \textbf{\underline{0.7248}} & 0.8033 & 0.7841 & \textbf{\underline{0.4464}} & \textbf{0.4073} & \textbf{0.9458} & \textbf{0.9561} \\
        AE & 0.7077 & 0.71 & 0.3685 & 0.255 & 0.5843 & 0.5551 & 0.7522 & 0.719 & 0.8076 & 0.7882 & 0.2233 & 0.1881 & \textbf{\underline{0.9439}} & \textbf{\underline{0.9513}} \\
        EncDecAD & 0.6744 & 0.6434 & 0.394 & 0.2602 & 0.5346 & 0.5328 & 0.6641 & 0.6271 & 0.5751 & 0.5277 & 0.2064 & 0.1723 & 0.9348 & 0.9256 \\
        Donut & 0.5618 & 0.5262 & 0.4014 & 0.2859 & 0.6296 & 0.617 & 0.6317 & 0.5765 & 0.6423 & 0.6089 & 0.1869 & 0.1556 & 0.5618 & 0.4383 \\
        FCVAE & 0.7389 & 0.7256 & \textbf{0.5991} & \textbf{0.4845} & 0.6281 & 0.6117 & 0.6838 & 0.6372 & 0.847 & 0.8242 & 0.3249 & 0.2813 & 0.7439 & 0.6557 \\
        SRCNN & 0.6253 & 0.5748 & 0.3072 & 0.2383 & 0.1816 & 0.1069 & 0.6267 & 0.5744 & 0.1605 & 0.0978 & 0.2358 & 0.2019 & 0.5446 & 0.5189 \\
        AnomalyTransformer & 0.306 & 0.1596 & 0.3655 & 0.2657 & 0.1331 & 0.0674 & 0.4326 & 0.2876 & 0.1751 & 0.0978 & 0.1243 & 0.0948 & 0.6075 & 0.4891 \\
        TFAD & 0.4578 & 0.362 & 0.15 & 0.0849 & 0.4658 & 0.3988 & 0.6165 & 0.569 & \textbf{\underline{0.9281}} & \textbf{\underline{0.9144}} & 0.4206 & 0.4034 & 0.3133 & 0.1913 \\
        TranAD & 0.6302 & 0.5726 & 0.4406 & 0.3254 & 0.4199 & 0.3944 & 0.588 & 0.5219 & 0.511 & 0.4551 & 0.1472 & 0.1261 & 0.8932 & 0.8696 \\
        TimesNet & 0.4663 & 0.4199 & 0.433 & 0.3105 & 0.3792 & 0.3232 & 0.4063 & 0.3349 & 0.5242 & 0.4988 & 0.0704 & 0.0501 & 0.9028 & 0.9174 \\
        OFA & 0.4689 & 0.3818 & 0.4145 & 0.2901 & 0.4664 & 0.4147 & 0.605 & 0.5593 & 0.7267 & 0.6979 & 0.1718 & 0.1371 & 0.7155 & 0.6536 \\
        FITS & 0.6623 & 0.6474 & 0.4341 & 0.302 & 0.5232 & 0.5423 & 0.6848 & 0.6573 & 0.9013 & 0.8903 & 0.2802 & 0.2585 & 0.8335 & 0.8122 \\
    \hline
    \end{tabular}
    }
    \label{tb:delay}
\end{table*}

\subsection{Trade-off Between Performance, Cost, and Efficiency} 
\begin{figure}[htb]
    \centering
    \includegraphics[width=0.45\textwidth]{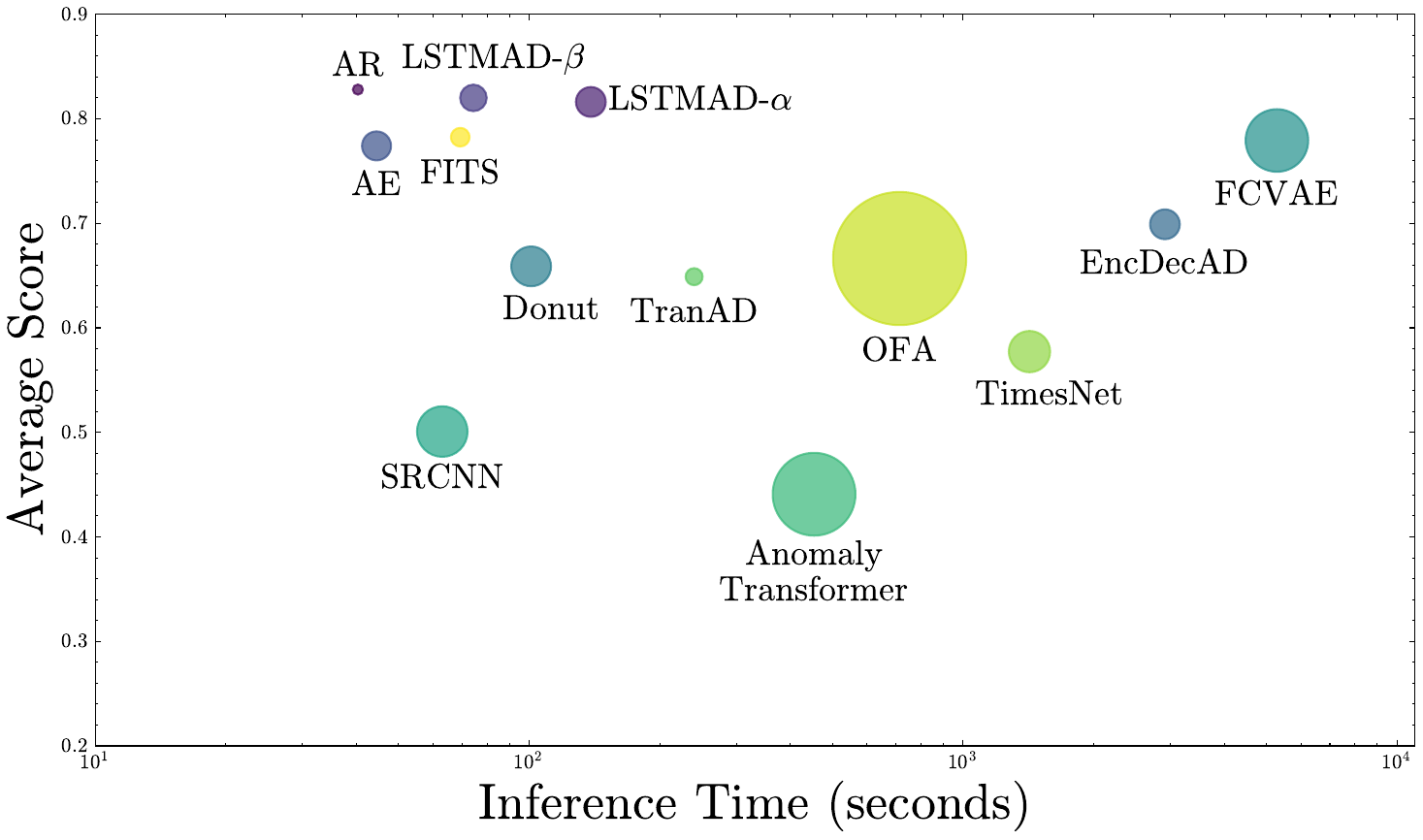}
    \caption{The triad of the model's performance, cost, and efficiency trade-off. The x-axis represents the total inference time (batch size set to 1) of the detectors on AIOPS\_7103fa0f-cac4-314f-addc-866190247439  (around 140,000 samples) under default parameters. The y-axis represents the average performance of the models on all datasets under the all-in-one learning schema. The size of the scatter points denotes each method's cube root of the number of parameters.} 
    \label{fig.trade-off}
\end{figure}

When deploying algorithms in real-world scenarios, it is often necessary to strike a balance between performance, storage costs, and inference speed. Notably different from other surveys, in this benchmark, we prefer inference time over training time due to the practical needs of anomaly detection. Each factor plays a critical role in determining the feasibility and practicality of the deployed algorithm. The trade-offs between these considerations are important to ensure an efficient and effective deployment that meets the specific requirements and constraints of the application. For example, real-time monitoring or critical systems may prioritize efficiency to detect anomalies promptly and respond quickly, while IoT (Internet of Things) devices with limited on-chip memory have to focus on the performance of models with a small number of parameters. We exclude the statistical methods, as their inference time is not a fixed value (polynomially correlated with the volume of historical data). Also, we exclude TFAD because it does not align with the real-time manner. As shown in Fig.~\ref{fig.trade-off}, the inference time for a single sample is far less than 50 milliseconds for all methods under our experimental platform. The learning-based AR exhibits the most favorable gain-to-cost ratio among all methods. FCVAE takes the longest time to inference due to its multi-epoch MCMC process. EncDecAD has the second longest inference time because it utilizes LSTM to perform inference for 100 time steps. In addition to providing guidance for practical applications, we aim for this perspective and toolkit to assist in discovering the scaling law in the field of time series anomaly detection under more reasonable inductive bias and larger parameter spaces \textbf{(RI \the\value{RI})}\stepcounter{RI}. 

\section{TimeSeriesBench Toolkit}

One of our primary intentions is to develop a suite that liberates practitioners from burdensome workflows, allowing them to engage in TSAD tasks with greater ease and efficiency. Existing suites narrowly offer convenience for the implementation of new methods with primitive workflows and rudimentary evaluations. Here, benefiting from the modular architecture of TimeSeriesBench, we provide a diverse range of extension interfaces implemented in Python to enable the community to conduct comparative experiments more flexibly and effortlessly, meanwhile reserving the possibilities for innovative and exceptionally demanding experiments, which might  unlock unexplored avenues of research.

\begin{figure}[htbp]
    \centering
    \includegraphics[width=0.45\textwidth]{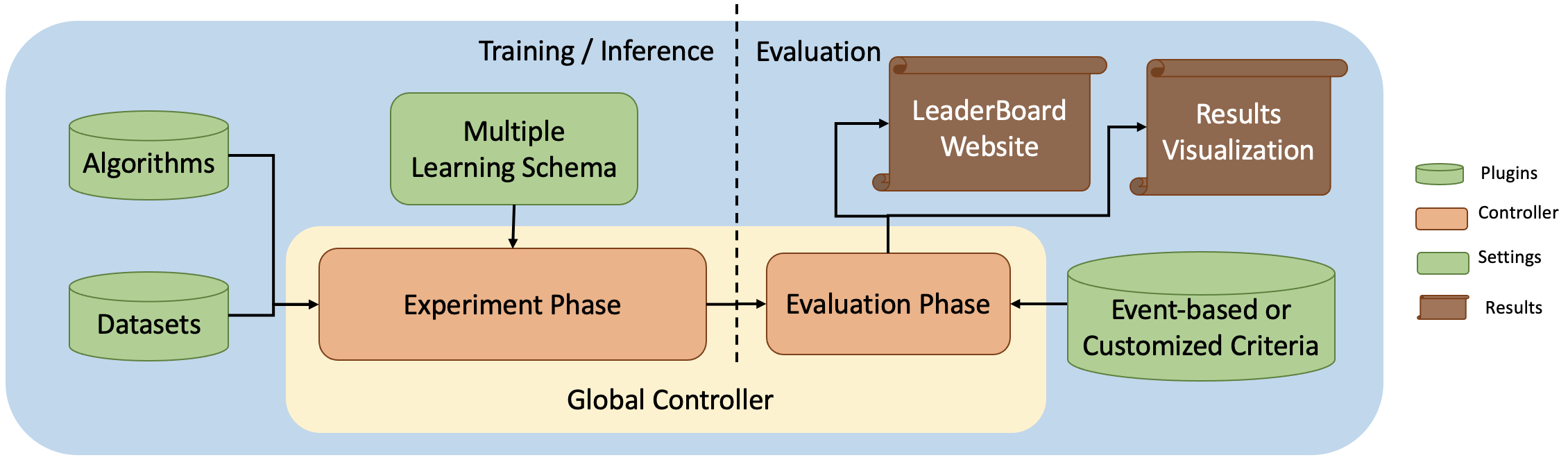}
    \caption{Overview of TimeSeriesBench Toolkit.}
    \label{fig:tsb}
\end{figure}

The overview of the toolkit is shown in Fig.~\ref{fig:tsb}. 
Compared to existing benchmark suites on univariate time series anomaly detection~\cite{suite_exa, suite_tods, suite_tsb, timeeval}, our framework provides more possibilities for diverse experiment setups. As shown in Table~\ref{tb:suite}, in addition to including rich datasets and diverse learning schemas, the toolkit considers the need for more kinds of evaluation criteria. We provide more famous evaluation criteria as built-in criteria. Thus researchers can conveniently compare the strengths and weaknesses of different criteria.

Despite the built-in settings, TimeSeriesBench toolkit also takes the requirements for extensibility into consideration. We provide flexible interfaces for dataset, method, evaluation criteria, runtime statistic (RT), and learning schema, while existing benchmarks only provides only parts of these interfaces. The dataset interface and algorithm interfaces are provided to allow evaluations on new or private datasets and novel methods under all learning schemas. All plots of raw curves and scores are saved for specific investigation. Previously mentioned flaws in evaluation criteria have sparked a research fervor and several latest works are dedicated to providing  evaluation criteria according to different assumptions~\cite{evaflaw3aaai, timesead, eval1, eval2vus, evaflaw1tisat, eval3}. Thus we also expose an interface for swiftly developing neo-criteria for evaluation grounded in realistic assumptions and evaluating the performance of all methods w.r.t various datasets. In some scenarios, there might be a trade-off between model accuracy and model training/testing cost. Therefore, we also provide a runtime statistics interface to conveniently track and analyze the runtime information of the model. Also, you can design your own training, test and other dataflows via implementing learning schema interface.

Owing to the flexible framework, our benchmark suite can incorporate real-world scenarios and challenges, aiming to replicate the complexities and nuances encountered in practical applications.
This enables researchers and practitioners to gain deeper insights into the strengths and weaknesses of various deep learning approaches, and is well-prepared for the emergence and development of Foundation Models (FMs) in the field of time series anomaly detection. In the toolkit, we provide four kinds of workflows to meet different research needs for time series anomaly detection: algorithms benchmarking, algorithm development, evaluation criteria development, and performance analysis. For more information about TimeSeriesBench, please refer to our repository and the website of the leaderboard.

% \usepackage{multirow}
\begin{table}[ht]
\centering
    \caption{Comparison among TimeSeriesBench and existing UTS anomaly detection benchmark suites. TimeSeriesBench presents more angles for performance evaluation, meanwhile offering user-friendly interfaces for up-to-date model and evaluation criterion developments.}
\resizebox{\linewidth}{!}{
\begin{tabular}{cc|ccccc}
\hline
\multicolumn{2}{c|}{\multirow{2}{*}{Suites}}                                                                                                                             & \multirow{2}{*}{Exathlon\cite{suite_exa}} & \multirow{2}{*}{TODS\cite{suite_tods}} & \multirow{2}{*}{TSB\cite{suite_tsb}} & \multirow{2}{*}{TimeEval\cite{timeeval}} & \multirow{2}{*}{Ours} \\
\multicolumn{2}{c|}{}                                                                                                                                                    &                           &                       &                      &     &                  \\ \hline
\multicolumn{1}{c|}{\multirow{2}{*}{\begin{tabular}[c]{@{}c@{}}Data\\ Source\end{tabular}}}              & Real-World                                                    & \ding{51}      & \ding{51}  & \ding{51}  & \ding{51} & \ding{51}  \\ % \cline{2-6} 
\multicolumn{1}{c|}{}                                                                                    & Synthetic                                                      & \ding{55}       & \ding{51}  & \ding{51}  & \ding{51} & \ding{51}  \\ \hline
\multicolumn{1}{c|}{\multirow{3}{*}{\begin{tabular}[c]{@{}c@{}}Learning\\ Schema\end{tabular}}}          & One-by-one                                                    & \ding{51}      & \ding{51}  & \ding{51}  & \ding{51} & \ding{51}  \\ % \cline{2-6} 
\multicolumn{1}{c|}{}                                                                                    & All-in-one                                                    & \ding{55}       & \ding{55}   & \ding{55}  & \ding{55} & \ding{51}  \\ % \cline{2-6} 
\multicolumn{1}{c|}{}                                                                                    & Zero-shot                                                     & \ding{55}       & \ding{55}   & \ding{55}  & \ding{55} & \ding{51}  \\ \hline
\multicolumn{1}{c|}{\multirow{3}{*}{\begin{tabular}[c]{@{}c@{}}Supported\\ Eval criteria\end{tabular}}} & Point-based\footnote{Point-based criteria treat each individual data point as a separate sample, disregarding the holistic characteristics of the anomaly segment.}                                                   & \ding{51} & \ding{51}  & \ding{51}  & \ding{51} & \ding{51} \\ % \cline{2-6} 
\multicolumn{1}{c|}{}                                                                                    & Range-based\footnote{Range-based criteria incorporate segment-level features, for instance, the detection latency, into the evaluation.}                                                   & \ding{51}      &    \ding{55}  & \ding{51}  & \ding{51} & \ding{51} \\ % \cline{2-6} 
\multicolumn{1}{c|}{}                                                                                    & Event-based                                               &     \ding{55}     &   \ding{55}   &   \ding{55}  & \ding{55}  & \ding{51}  \\ \hline
\multicolumn{1}{c|}{\multirow{5}{*}{Extensibility}}   
& Dataset                                                        & \ding{55}       & \ding{51}  & \ding{51}   & \ding{51} & \ding{51}  \\ % \cline{2-6}
\multicolumn{1}{c|}{} 
& Method                                                        & \ding{51}       & \ding{51}  & \ding{55}   & \ding{51} & \ding{51}  \\ % \cline{2-6} 
\multicolumn{1}{c|}{}                                                                                    & Eval criteria & \ding{55}       & \ding{55}   & \ding{55}   & \ding{51} & \ding{51}  \\ 
\multicolumn{1}{c|}{}                                                                                    & RT Statistics & \ding{55}       & \ding{55}   & \ding{55}   & \ding{55} & \ding{51}  \\
\multicolumn{1}{c|}{}   
& Learning Schema & \ding{55}       & \ding{55}   & \ding{55}   & \ding{55} & \ding{51}  \\
\hline
\end{tabular}
}
\label{tb:suite}
\end{table}
 
% The following are typical audiences for TimeSeriesBench toolkit:
% \begin{itemize}
%     \item \textit{Algorithm Researchers}. TimeSeriesBench offers a flexible interface for implementing, training, and testing new algorithms. Researchers can choose all the mentioned schemas for model training and testing. The framework also provides a full pipeline for loading datasets, running experiments, conducting evaluations, and performing analysis such as generating plots and comparing anomaly scores.
%     \item \textit{Evaluation Researchers}. TimeSeriesBench offers a flexible interface for implementing evaluation criteria based on anomaly scores and ground truth labels. Researchers can easily evaluate existing methods according to their specific criteria. The framework also allows evaluation based on offline scores of methods generated by merely once training and test phase.
%     \item \textit{Practitioners of Community or Enterprise}. TimeSeriesBench provides a unified and clear dataset format, making it easy to introduce private datasets. It also allows for easy performance comparison of baselines on custom datasets, providing overall performance in CSV format based on criteria suitable for specific applications. Moreover, TimeSeriesBench allows practitioners to record runtime statistics such as model parameter size and inference time, enabling trade-offs between performance, cost, and efficiency.
% \end{itemize}

\section{Threats to validity}
As a benchmark, our research is susceptible to several common threats that can compromise the validity and reliability of our findings, including dataset, algorithm settings and the applicable scenarios.

\noindent\textbf{Dataset.} Although we have adopted six well-known datasets and have released a real-world dataset, the scenarios covered by these datasets are still limited, so the evaluation results in other scenarios may vary from the main results. We will supplement more datasets in the future to make the evaluation results more generalizable.

\noindent\textbf{Algorithms settings.} Due to time and computational resource constraints, we evaluated the algorithms using their default parameter settings in the experiments. If the hyperparameters are fine-tuned for each scenario, some methods might achieve better performance.

\noindent\textbf{Applicable scenarios.} In this paper we focus on \textit{real-time} anomaly detection scenarios, and introduce a more robust criterion for evaluation based on real-time detection demands. All results and analysis are based on this setting. However, for offline detection tasks that do not have real-time requirements and require precise detection of the duration of anomalies, other evaluation criteria such as VUS~\cite{eval2vus} may be more suitable. You can specify this criterion as the evaluation criteria in TimeSeriesBench toolkit to obtain results that are more tailored to your specific application scenario.

\section{Conclusion AND FUTURE DIRECTIONS}
In this paper, we propose TimeSeriesBench, a comprehensive and application-oriented benchmark for evaluating the performance of existing and emerging UTS anomaly detection methods. TimeSeriesBench takes into account existing industrial concerns and conducts a comprehensive performance evaluation of some well-known and latest methods under settings that meet industrial requirements. Also, it offers unprecedented perspectives for measuring algorithm performance, meanwhile laying a solid foundation for the development of Foundation Models in the field. Moreover, TimeSeriesBench includes a user-friendly toolkit and leaderboard, offering easy-to-use interfaces that allow researchers and practitioners to focus on advancing their algorithms without getting entangled in repetitive tasks. We intend to incorporate more latest methods/evaluation criteria and employ more high-quality data. Also, we will continue to monitor the latest developments in general foundation models for time series within the realm of anomaly detection, adapting them accordingly within our benchmark. Additionally, whether you propose deep learning or statistical methods, we welcome your participation in our leaderboard to help drive the development of the time series anomaly detection community.

\section*{ACKNOWLEDGMENTS}
% This work is supported by the National Key R\&D Program of China (2021YFE0111500), the Chinese Academy of Sciences (No.241711KYSB20200023), and the National Natural Science Foundation of China (No.62202445), in part by the State Key Program of National Natural Science Foundation of China under Grant 62321166652.
This work was supported by the National Key Research
and Development Program of China (No. 2021YFE0111500),
in part by the Chinese Academy of Sciences (No.
241711KYSB20200023), in part by the National Natural Sci-
ence Foundation of China (No. 62202445), and in part by the
State Key Program of the National Natural Science Foundation
of China under Grant 62321166652.

\bibliographystyle{ieeetr}
\bibliography{main}

\end{document}